\documentclass[a4paper,num-refs]{mod}
\usepackage{graphicx}
\usepackage{float}
\usepackage{siunitx}

\usepackage{subcaption}
\usepackage{adjustbox}
\usepackage{placeins} 

\title{Segmentation of Roots in Soil with U-Net}

\author[1,\authfn{1}]{Abraham George Smith}
\author[2]{Jens Petersen}
\author[2]{Raghavendra Selvan}
\author[1]{Camilla Ruø Rasmussen}

\affil[1]{Department of Plant and Environmental Sciences, University of Copenhagen}
\affil[2]{Department of Computer Science, University of Copenhagen}

\authnote{\authfn{1}ags@plen.ku.dk}

\papercat{Paper}

\runningauthor{Smith et al.}
\runningtitle{Segmentation of Roots in Soil with U-Net}
\begin{document}

\begin{frontmatter}
\date{*ags@plen.ku.dk}
\maketitle
\begin{abstract}

Plant root research can provide a way to attain stress-tolerant crops that
produce greater yield in a diverse array of conditions. Phenotyping roots in soil is 
often challenging due to the roots being difficult to access ad the use of time consuming manual methods. 
Rhizotrons allow visual inspection of root growth through transparent surfaces.
Agronomists currently manually label photographs of roots obtained from rhizotrons using a line-intersect method to obtain root length
density and rooting depth measurements which are essential for their
experiments.

We investigate the effectiveness of an automated image segmentation method based on the U-Net Convolutional Neural Network (CNN) architecture to enable such measurements.  We design a data-set of 50 annotated Chicory 
(Cichorium intybus L.) root images which we use to train, validate and test the system and compare against a baseline built using the Frangi vesselness filter. We obtain metrics using manual annotations and line-intersect counts. 

Our results on the held out data show our proposed automated segmentation
system to be a viable solution for detecting and quantifying roots. We evaluate
our system using 867 images for which we have obtained line-intersect counts,
attaining a Spearman rank correlation of 0.9748 %$(p < 10^{-300})$
and an $r^2$ of 0.9217. We also achieve an $F_1$ of 0.7 when comparing the automated
segmentation to the manual annotations, with our automated segmentation system
producing segmentations with higher quality than the manual annotations for
large portions of the image.

\end{abstract}


\end{frontmatter}

\section{Background}
\label{sec:background}
High-throughput phenotyping of roots in soil has been a long-wished-for goal for various
research purposes \citep{trachsel2011shovelomics, Wasson2012, Wasson2016,
maeght2013study}. The challenge of exposing the architecture of roots hidden in soil has
promoted studies of roots in artificial growth media \citep{Lynch2012}. However, root
growth is highly influenced by physical constraints \citep{Bengough1991} and such studies
have shown to be unrepresentative of roots in soil \citep{Wojciechowski2009, Watt2013}.

Traditionally studies of roots in soil have relied on destructive and laborious
methods such as trenches in the field and soil coring followed by root washing
\citep{Noordwijk1979}. Recently 3D methods such as X-ray computed tomography
\citep{Mooney2011} and magnetic resonance imaging \citep{Stingaciu2013} have been
introduced, but these methods require expensive equipment and only allow small samples.

Since the 1990, rhizotrons \citep{taylor1970measurement, Huck1982, Vandegeijn1994} and
minirhizotrons \citep{Taylor1990, Johnson2001} which allow non-invasive monitoring of
spatial and temporal variations in root growth in soil, have gained popularity.
Minirhizotrons facilitate the repeated observation and photographing of roots through
the transparent surfaces of below ground observation tubes \citep{rewald2013minirhizotron}.

A major bottleneck when using rhizotron methods is the extraction of relevant information
from the captured images. Images have traditionally been annotated manually using the
line-intersect method where the number of roots crossing a line in a grid is counted and
correlated to total root length \citep{newman1966method, tennant1975test} or normalised
to the total length of grid line
\citep{thorup2001differences}. The line-intersect method was originally developed for
washed roots but is now also used in rhizotron studies where a grid is either directly
superimposed on the soil-rhizotron interface \citep{Upchurch1983, Ulas2012} or indirectly
on recorded images \citep{Heeraman1993, Thorup-Kristensen2015}. The technique is arduous
and has been reported to take 20 minutes per metre of grid line in minirhizotron studies
\citep{Boehm1977}. Line-intersect counts are not a direct measurement of root length and
do not provide any information on architectural root traits such as branching, diameter,
tip count, growth speed or growth angle of laterals.

To overcome these issues, several attempts have been made to automate the detection and
measurement of roots, but all of them require manual supervision, such as mouse clicks to
detect objects \citep{lobet2013online, cai2015rootgraph}. 

The widely used ``RootFly'' software provides both manual annotation and automatic root detection functionality \citep{Zeng2007}.
Although the automatic detection worked well on the initial three datasets the authors found it did not transfer well to new soil types (personal communication with Stan Birchfield, September 27, 2018).

Following the same manual annotation procedure as in RootFly, \cite{Ingram2001} calculated that it takes 1--1.5 hours per 100 $cm^2$ to annotate images of roots from minirhizotrons, adding up to thousands of hours for many minirhizotron experiments. Although existing software is capable of attaining much of the desired information, the annotation time required is prohibitive and severely limits the use of such tools.

Image segmentation is the splitting of an image into different meaningful parts. A fully automatic root segmentation system would not just save agronomists time but could also provide more localized information on which roots have grown and by how much as well as root width and architecture.

The low contrast between roots and soil has been a challenge in previous attempts to automate root detection. Often only young unpigmented roots can be detected \citep{Vamerali1999} or roots in black peat soil \citep{Nagel2012}. To enable detection of roots of all ages in heterogeneous field soils, attempts have been made to increase the contrast between soil and roots using custom spectroscopy. UV light can cause some living roots to fluoresce and thereby stand out more clearly \citep{Wasson2016} and light in the near–infrared spectrum can increase the contrast between roots and soil \citep{Nakaji2007}.

Other custom spectroscopy approaches have shown the potential to distinguish between living and dead roots \citep{wang1995accuracy, Smit1996} and roots from different species
\citep{Goodwin1948, Rewald2012}. A disadvantage of such approaches is that they require
more complex hardware which is often customized to a specific experimental setup. A method
which works with ordinary RGB photographs would be attractive as it would not require
modifications to existing camera and lighting setups, making it more broadly applicable to
the wider root research community. Thus in this work we focus on solving the problem of
segmenting roots from soil using a software driven approach. 

Prior work on segmenting roots from soil in photographs has used feature extraction
combined with traditional machine learning methods \citep{zeng2006detecting,
zeng2010rapid}.  A feature extractor is a function which transforms raw data into a
suitable internal representation from which a learning subsystem can detect or classify patterns \citep{lecun2015deep}.  The process of manually designing a feature extractor is
known as feature engineering.  Effective feature engineering for plant phenotyping
requires a practitioner with a broad skill-set as they must have sufficient knowledge of
both image analysis, machine learning and plant physiology \citep{tsaftaris2016machine}. Not
only is it difficult to find the optimal description of the data but the features found
may limit the performance of the system to specific datasets \citep{pound2017deep}.  With
feature engineering approaches, domain knowledge is expressed in the feature extraction
code so further programming is required to re-purpose the system to new datasets. 

Deep learning is a machine learning approach, conditioned on the training procedure, where a machine fed with raw data
automatically discovers a hierarchy of representations that can be useful for detection or classification tasks \citep{lecun2015deep}. Convolutional Neural Networks (CNNs) are a
class of deep learning architectures where the feature extraction mechanism is encoded in the weights (parameters) of the network, which can be updated without the need for manual
programming by changing or adding to the training data. Via the training process a CNN is able to learn from examples, to approximate the labels or annotations for a given input. This makes the 
effectiveness of CNNs highly dependent on the quality and quantity of the annotations provided.

Deep learning facilitates a decoupling of plant physiology domain knowledge and machine learning technical expertise. A deep learning practitioner can focus on the selection and optimisation of a general purpose neural network architecture whilst root experts encode their domain knowledge into annotated data-sets created using image editing software.

CNNs have now established their dominance on almost all
recognition and detection tasks \citep{krizhevsky2012imagenet, farabet2013learning,
szegedy2015going, tompson2015efficient}. They have also been used to segment roots from
soil in X-ray tomography \citep{douarre2016deep} and to identify the tips of wheat roots
grown in germination paper growth pouches \citep{pound2017deep}. CNNs have an
ability to transfer well from one task to another, requiring less training data for new
tasks \citep{tajbakhsh2016convolutional}. This gives us confidence that knowledge attained
from training on images of roots in soil in one specific setup can be transferred to a new
setup with a different soil, plant species or lighting setup.

The aim of this study is to develop an effective root segmentation system using a CNN. We
use the U-Net CNN architecture \citep{ronneberger2015u}, which has proven to be especially
useful in contexts where attaining large amounts of manually annotated data is challenging,
which is the case in biomedical or biology experiments. 

As a baseline machine learning approach we used the Frangi vessel enhancement filter
\citep{frangi1998multiscale}, which was originally developed to enhance vessel structures
on images of human vasculature. Frangi filtering represents a more traditional and simpler off-the-shelf approach which
typically has lower minimum hardware requirements and training time when compared to
U-Net. 

We hypothesize that (1) U-Net will be able to effectively discriminate between roots and soil in RGB photographs, demonstrated by a strong positive correlation between root length density obtained from U-Net segmentations and root intensity obtained
from the manual line-intersect method. And (2) U-Net will outperform a traditional machine learning approach with larger amounts of agreement between the U-Net segmentation output and the test set annotations.
\section{Data collection}
\FloatBarrier % don't let figures float outside their section

\begin{table}
    \centering
\caption{Number of images from each date. Not all images are included as they may contain large amounts of equipment. 
        }
\label{tab:imcounts}
\begin{tabular}{llll}
Date & Total images & Included & Line-intersect counts \\
    \toprule
    21/06/16 & 192 & 168 & Yes \\
    27/06/16 & 296 & 180 & No \\
    04/07/16 & 320 & 196 & Yes \\
    11/07/16 & 348 & 216 & No \\
    18/07/16 & 396 & 248 & Yes \\
    25/07/16 & 420 & 268 & No \\
    22/08/16 & 440 & 280 & Yes \\
    05/09/16 & 440 & 276 & No \\
    21/09/16 & 448 & 280 & No \\
\end{tabular}
\end{table}
\begin{figure}
    \centering
    \includegraphics[width=1.0\linewidth]{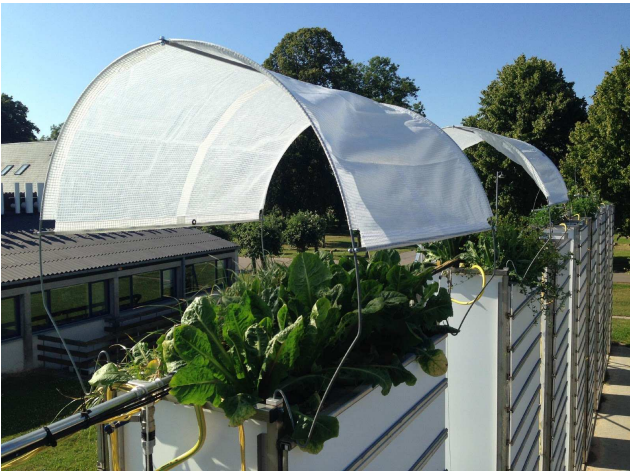}
    \caption{Chicory (Cichorium intybus L.) shown from above in the rhizotron facility.}
     \label{fig:chicory_02}
\end{figure}

We used images of chicory (Cichorium intybus L.) taken during summer 2016 from a large 4 m
deep rhizotron facility at University of Copenhagen, Taastrup, Denmark (Figure \ref{fig:chicory_02}). The images had been used in a previous study \cite{rasmussen2018chicory} where the analysis was performed using the manual line-intersect method. As we make no modifications to the hardware or photographic procedure, we are able to evaluate our method as a drop-in replacement to the manual line-intersect method.

The facility from which the images were captured consists of 12 rhizotrons. Each rhizotron is
a soil filled rectangular box with 20 1.2 m wide vertically stacked transparent acrylic
panels on two of its sides which are covered by 10 mm foamed PVC plates. These plates can be removed to allow inspection of root growth at
the soil-rhizotron interface. 
% Each panel covers the full width of a rhizotron. % Camilla says not nessaery.
There were a total of 3300 images which had been taken on
9 different dates during 2016. The photos were taken from depths between 0.3 and 4 m. Four
photos were taken of each panel in order to cover its full width, with each individual
image covering the full height and 1/4 of the width (For further details of the experiment and the facility see  \cite{rasmussen2018chicory}). The image files were labelled
according to the specific rhizotron, direction and panel they are taken from with the
shallowest which is assigned the number 1 and the deepest panel being assigned the number 20.

Line-intersect counts were available for 892 images. They had been obtained using a version of the
line-intersect method \citep{newman1966method} which had been modified to use grid
lines \cite{marsh1971measurement, tennant1975test} overlaid over an image to compute root
intensity. Root intensity is the number of root intersections per metre of grid line in each
panel \cite{thorup2001differences}. 

In total four different grids were used. Coarser grids were used
to save time when counting the upper panels with high root intensity and finer grids were used to ensure low variation in counts from the lower panels with low root intensity. The 4 grids used had squares of sizes 10, 20, 40 and 80 mm. The grid size for each depth was selected by the counter, aiming to have at least 50 intersections for all images obtained from that depth. For the deeper panels with less roots, it was not possible to obtain 50 intersections per panel so the finest grid (10 mm) was always used.

We only used photos that had been deemed suitable for analysis by the manual
line-intersect method. From the 3300 originals, images from panels 3, 6, 9, 12, 15 and
18 were excluded as they contained large amounts of equipment such as cables
and ingrowth cores. Images from panel 1 were excluded as it was not fully covered with soil. 
%Line-intersect counts were available for images taken on 4 out of the 9 dates.
Table \ref{tab:imcounts} shows the number of images from each date, the number of images remaining after excluding panels unsuitable for analysis and if line-intersect counts were available.  

\begin{figure}
     \centering
     \includegraphics[width=1.00\linewidth]{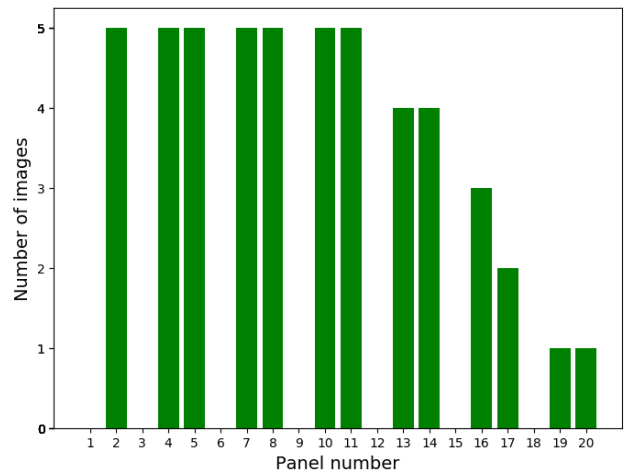} 
     \caption{The number of images selected for annotation from each panel depth. \label{fig:annot_panel_dist}}
\end{figure}

Deeper panels were sometimes not photographed as when photographing the panels the
photographer worked from the top to the bottom and stopped when it was clear that no deeper roots could be observed. We took the depth distribution of all images obtained from the rhizotrons in 2016 into account when selecting images for annotation in order to create a representative sample (Figure \ref{fig:annot_panel_dist}).
After calculating how many images to select
from each depth the images were selected at random

The first 15 images were an exception
to this. They had been selected by the annotator whilst aiming to include all depths. We kept these images but ensured they were not used in the final evaluation of model performance as we were uncertain as to what biases had led to their selection.

\vspace{2mm}

\subsection{Annotation}
%Due to time constraints
We chose a total of 50 images for annotation. This number was based on the availability of
our annotator and the time requirements for annotation.

To facilitate comparison with the available root intensity measurements by analysing the same region of the image as \cite{rasmussen2018chicory}, the images were cropped from their original dimensions of $4608\times2592$ pixels to $3991\times1842$ pixels which corresponds to an area of approximately 300 $\times$ 170 mm of the surface of the rhizotron. This was done by removing the right side of the image where an overlap between images is often present and the top and bottom which included the metal frame around the acrylic glass.

\begin{figure*}
\vspace{1cm}
\begin{minipage}[t]{0.5\textwidth}
    %\begin{picture}(200, 200) % found by trial and error
    %\put(0,0){
       %\adjincludegraphics[width=\textwidth, trim={{0.7\width} {0.4\height} 0 0},clip]{images/method/16_07_18_11E8c_P7181734_photo.jpg}
       \includegraphics[width=\textwidth]{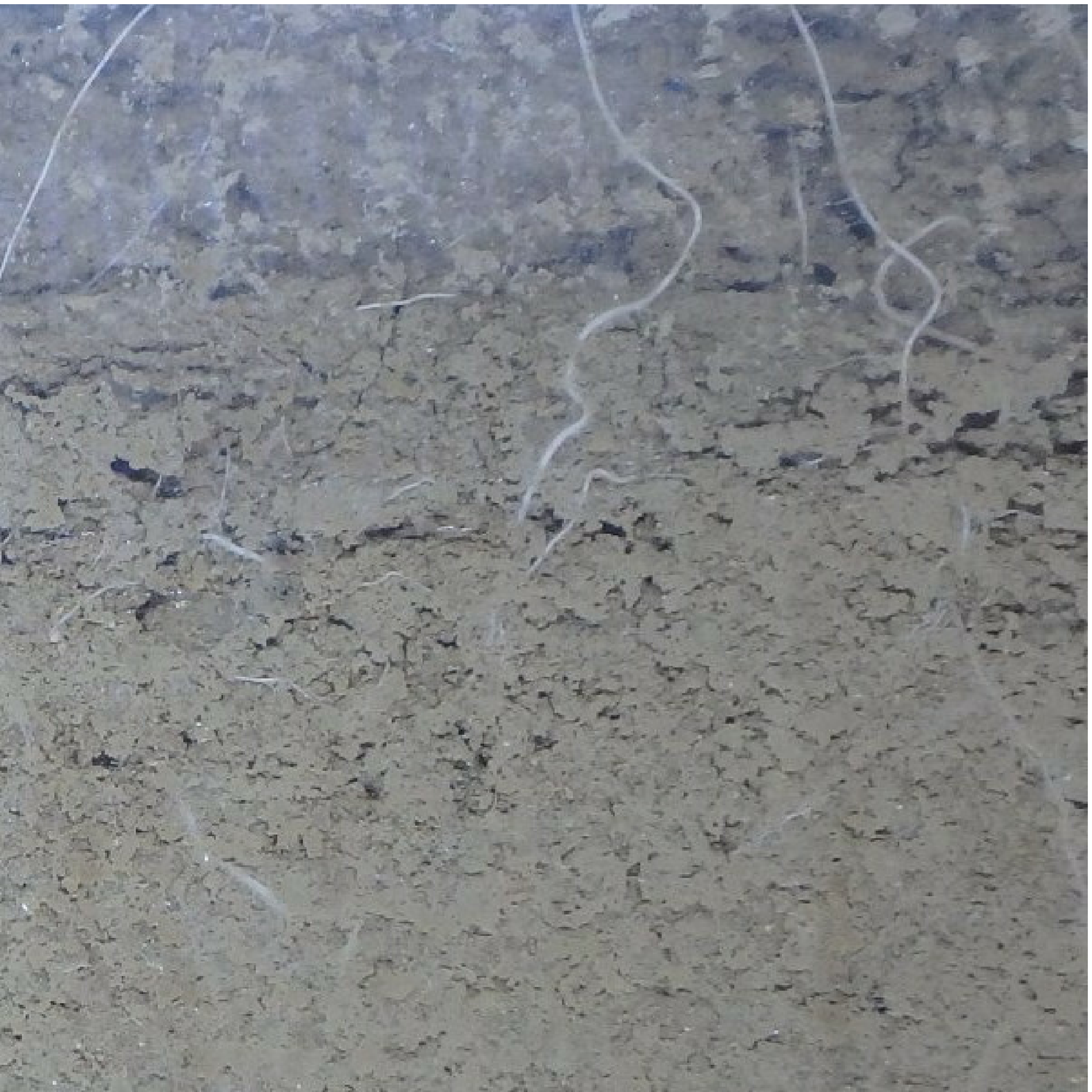}
    %}
    %\put(10, 10){\huge{a}} % 10, 200 is top left
    %\end{picture}
\end{minipage}
\begin{minipage}[t]{0.5\textwidth}
    %\begin{picture}(0.7\width, 0.4\height)
    %\put(0,0) {
    \includegraphics[width=\textwidth]{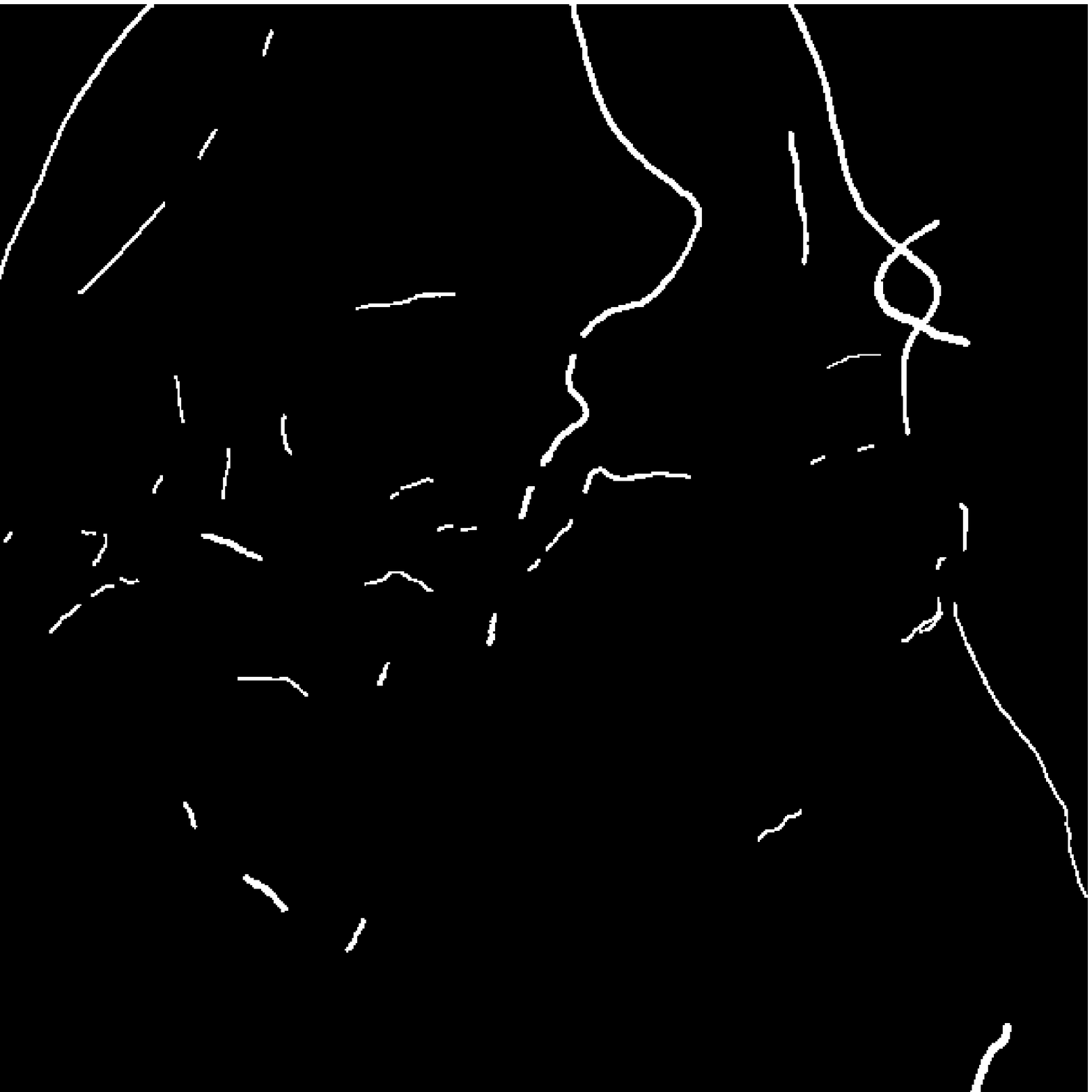}
    %}
    %\put(10, 10){\huge{\color{white} b}}
    %\end{picture0i}
\end{minipage}
\caption{Part of a photo shown with corresponding annotation.
    %%a) Sub region of one of the photos in the training data. 
    The photo (left) shows roots and soil as seen through the transparent acrylic glass on the surface of one of the rhizotrons and the annotation of the roots in the photo. The annotation (right) show root pixels in
    white and all other pixels in black. Annotations like these were used for training the U-Net CNN.
}\label{fig:photo_annot}

\end{figure*}

A detailed per-pixel annotation (Figure \ref{fig:photo_annot}) was then created as a separate layer in Photoshop by a trained agronomist with extensive experience using the line-intersect method. Annotation took approximately 30 minutes per image with the agronomist labelling all pixels which they perceived to be root. This resulted in 7351422 pixels labelled as either root or soil for each image.

%\begin{figure*}
%\vspace{1cm}
%\begin{minipage}[t]{0.5\textwidth}
    %\begin{picture}(200, 200) % found by trial and error
    %\put(0,0){
       %\adjincludegraphics[width=\textwidth, trim={{0.7\width} {0.4\height} 0 0},clip]{images/method/16_07_18_11E8c_P7181734_photo.jpg}
%       \includegraphics[width=\textwidth]{images/method/16_07_18_11E8c_P7181734_photo}
%    }
%    \put(10, 10){\huge{a}} % 10, 200 is top left
%    \end{picture}
%\end{minipage}
%\begin{minipage}[t]{0.5\textwidth}
%    \begin{picture}(0.7\width, 0.4\height)
%    \put(0,0) {
%    \includegraphics[width=\textwidth]{images/method/16_07_18_11E8c_P7181734_annotation}
%    }
%    \put(10, 10){\huge{\color{white} b}}
%    \end{picture}
%\end{minipage}
%\caption{Part of a photo shown with corresponding annotation.
%    (a) Sub region of one of the photos in the training data. The images shows roots and soil as seen through the transparent acrylic glass on the surface of one of the rhizotrons.
%    (b) Annotation of the roots in the photo. The annotations show root pixels in
%    white and all other pixels in black. Annotations like these were used for training the U-Net CNN.
%}\label{fig:photo_annot}
%\end{figure*}

\subsection{Data split}

During the typical training process of a neural network, the labelled or annotated data is split into a training, validation and test dataset.
The training set is used to optimize a neural network using a
process called Stochastic Gradient Descent (SGD) where the weights (parameters) are adjusted in such a way that segmentation performance improves. The validation set is used for giving an indication of system performance during the training procedure and tuning the so-called hyper-parameters, not optimised by SGD such as the learning rate. See the section U-Net Implementation for more details.
The test set performance is only calculated once after the neural network training process is complete to ensure an unbiased indication of performance.

We selected 10 images for the test set which meant the full range of panel heights could not be included. One image was selected from all panel heights except for 13, 17, 18 and 20. After setting aside the test set, we inspected the other images. We removed two images. One
because it didn't contain any roots and another because a sticker was present on top of the acrylic glass. The remaining 38 images were then split into training and validation datasets. We used the root pixel count from the annotations to guide the split of the images into a train and validation data-set. The images were ordered by the number of root pixels in each image and then 9 evenly spaced images were selected for the validation set with the rest being assigned to the training set. This was to ensure a range of root intensities was present in both training and validation sets.

\subsection{Metrics}

To evaluate the performance of the model during development and testing we used $F_1$. 
We selected $F_1$ as a metric because we were interested in a system which would
be just as likely to overestimate as it would underestimate the roots in a given
photo. That meant precision and recall were valued equally. In this context
precision is the ratio of correctly predicted root pixels to the number of pixels
predicted to be root and recall is the ratio of correctly predicted root pixels to the
number of actual root pixels in the image. Both recall and precision must be high for
$F_1$ to be high. 
\begin{equation}
    F_{1} = 2 \cdot {\frac{\text{precision} \cdot \text{recall}}
                         {\text{precision} + \text{recall}}}
\end{equation}
The $F_1$ of the segmentation output was calculated using the training and validation sets during system development. The completed system was then evaluated using the test set in order to provide a measure of performance on unseen data. We also report accuracy, defined as the ratio of correctly predicted to total pixels in an image.

In order to facilitate comparison and correlation with line-intersect counts,
we used an approach similar to \cite{Andren1996} to convert a root segmentation to a length estimate. The scikit-image skeletonize function was used to first thin the segmentation and then the remaining pixels were counted. This approach was used for both the baseline and the U-Net segmentations.

For the test set we also measured correlation between the root length of the output segmentation and the manual root intensity given by the
line-intersect method. We also measured correlation between the root length of our manual per-pixel annotations and the U-Net output segmentations for our held out test set. To further quantify the effectiveness of the system as a replacement for the
line-intersect method, we obtained the coefficient of determination ($r^2$) for the root length given by
our segmentations and root intensity given by the line-intersect method for 867 images. Although line-intersect counts were available for 892 images, 25 images were excluded from our correlation analysis as they had been used in the training dataset.

\subsection{Frangi Vesselness Implementation}

    For our baseline method we build a system using the Frangi Vesselness enhancement filter
    \citep{frangi1998multiscale}. We selected the Frangi filter based on the
    observation that the roots look similar in structure to blood vessels, for which the
    Frangi filter was originally designed. We implemented the system using the Python
    programming language (version 3.6.4), using the scikit-image \citep{van2014scikit}
    (version 0.14.0) version of Frangi. Vesselness refers to a measure of tubularity that is predicted by the Frangi filter for a given pixel in the image. To obtain a segmentation using the Frangi filter we thresholded the output so only regions of the image above a certain vesselness level
    would be classified as roots. To remove noise we further processed the segmentation output using connected component analysis to remove regions less than a threshold of connected pixels.
    To find optimal parameters for both the thresholds and
    the parameters for the Frangi filter we used the Covariance Matrix Adaptation
    Evolution Strategy (CMA-ES) \citep{hansen2016cma}. In our case the objective function
    to be minimized was $1 - mean(F_1)$ where $mean(F_1)$ is the mean of the $F_1$ scores
    of the segmentations produced from the thresholded Frangi filter output. 
    
\subsection{U-Net Implementation}
    % Architecture, loss, training schedule, augmentation.\\
    \subsubsection{Architecture}

    We implemented a U-Net CNN in Python (version 3.6.4) using
    PyTorch \citep{paszke2017automatic} which is an open source machine learning library
    which utilizes GPU accelerated tensor operations. PyTorch has convenient utilities for defining and
    optimizing neural networks. We used an NVIDIA TITAN Xp 12 GB GPU\@.
    Except for the input layer which was modified to
    receive RGB instead of a single channel, our network had the same number of layers and
    dimensions as the original U-Net \cite{ronneberger2015u}. We applied Group norm \citep{wu2018group} after all
    ReLU activations as opposed to Batch norm \citep{ioffe2015batch} as batch sizes as
    small as ours can cause issues due to inaccurate batch statistics degrading the
    quality of the resulting models \citep{ioffe2017batch}. The original U-Net proposed in
    \citep{ronneberger2015u} used Dropout which we avoided as in some cases the
    combination of dropout and batch normalisation can cause worse results
    \citep{li2018understanding}. He initialisation \citep{he2015delving} was used for all
    layers. 
    
    \begin{figure}
          \centering
          \includegraphics[width=\linewidth]{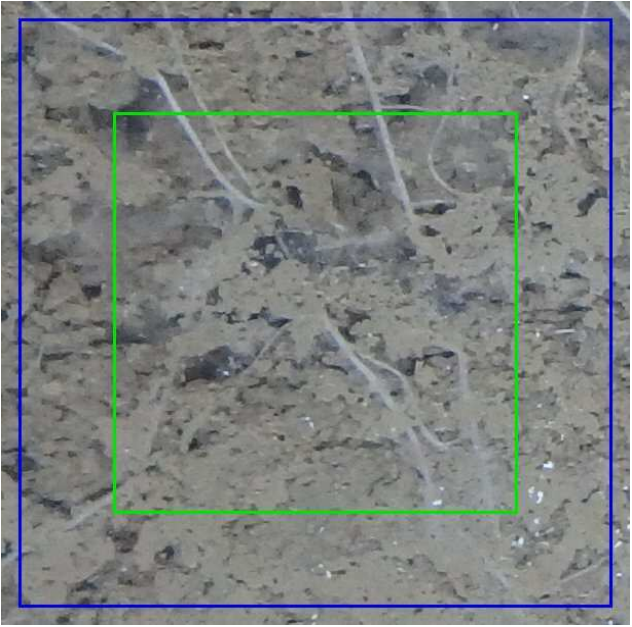}
            \caption{U-Net receptive field (input) size (blue) and output size (green).
                    The receptive field is the region of the input data which is provided 
                    to the neural network. The output size is the region of the original image which the output segmentation is for. The output is smaller than the input to ensure sufficient context for the classification of each pixel in the output.}
           \label{fig:u_patch}
          \centering
    \end{figure}

    \subsubsection{Instance selection}
    The network takes tiles with size $572 \times 572$ as input and outputs a segmentation
    for the centre $388 \times 388$ region for each tile (Figure \ref{fig:u_patch}).
    We used mirroring to pad the full image before extracting tiles. Mirroring in this context means the image was reflected at the edges to make it bigger and provide some synthetic context to allow segmentation at the edges of the image. In neural network
    training an epoch refers to a full pass over the training data.  Typically several
    epochs are required to reach good performance. At the start of each epoch we extracted
    90 tiles with random locations from each of the training images. These tiles were
    then filtered down to only those containing roots and then a maximum of 40 was taken from
    what ever was left over. This meant images with many roots would still be limited to
    40 tiles. The removal of parts of the image which does not contain roots has
    similarity to the work of \cite{kayalibay2017cnn} who made the class imbalance problem
    less severe by cropping regions containing empty space. When training U-Net with mini
    batch SGD, each item in a batch is an image tile and multiple tiles are input into
    the network simultaneously. Using tiles as opposed to full images gave us more
    flexibility during experimentation as we could adjust the batch size depending on the
    available GPU memory. When training the network we used a batch size of 4 to ensure we
    did not exceed the limits of the GPU memory. Validation metrics were still calculated
    using all tiles with and without soil in the validation set.
       
    \subsubsection{Preprocessing and augmentation}

\begin{figure}
        \centering
        % \begin{subfigure}[t]{0.75\linewidth}
        \begin{picture}(200, 200) % found by trial and error
        \put(10,0){
        \includegraphics[width=0.75\linewidth]{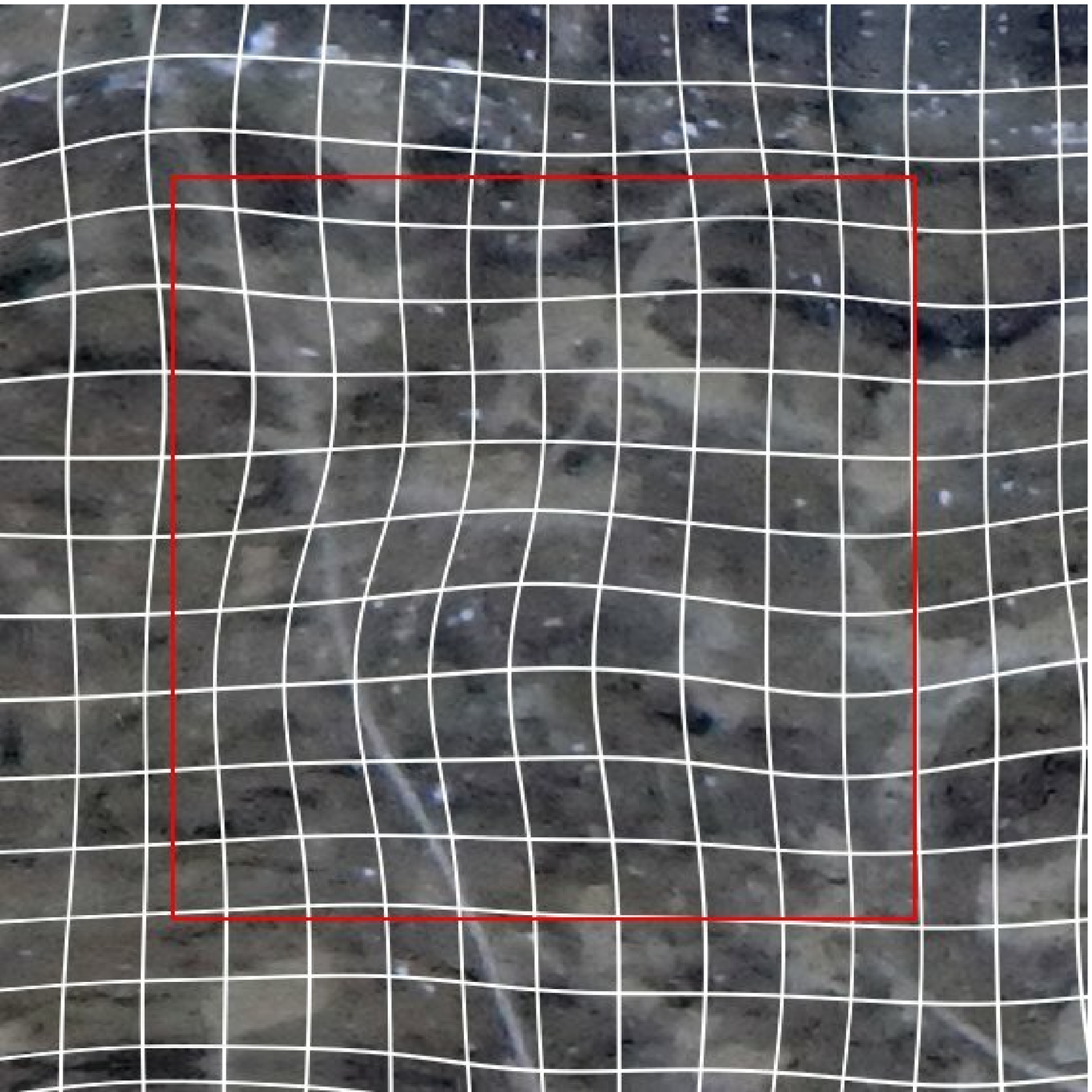}
        }
        \put(-3, 7){\huge{a}} % 10, 200 is top left
    \end{picture}
    \begin{picture}(200, 200) % found by trial and error
        \put(10,0){
        \includegraphics[width=0.75\linewidth]{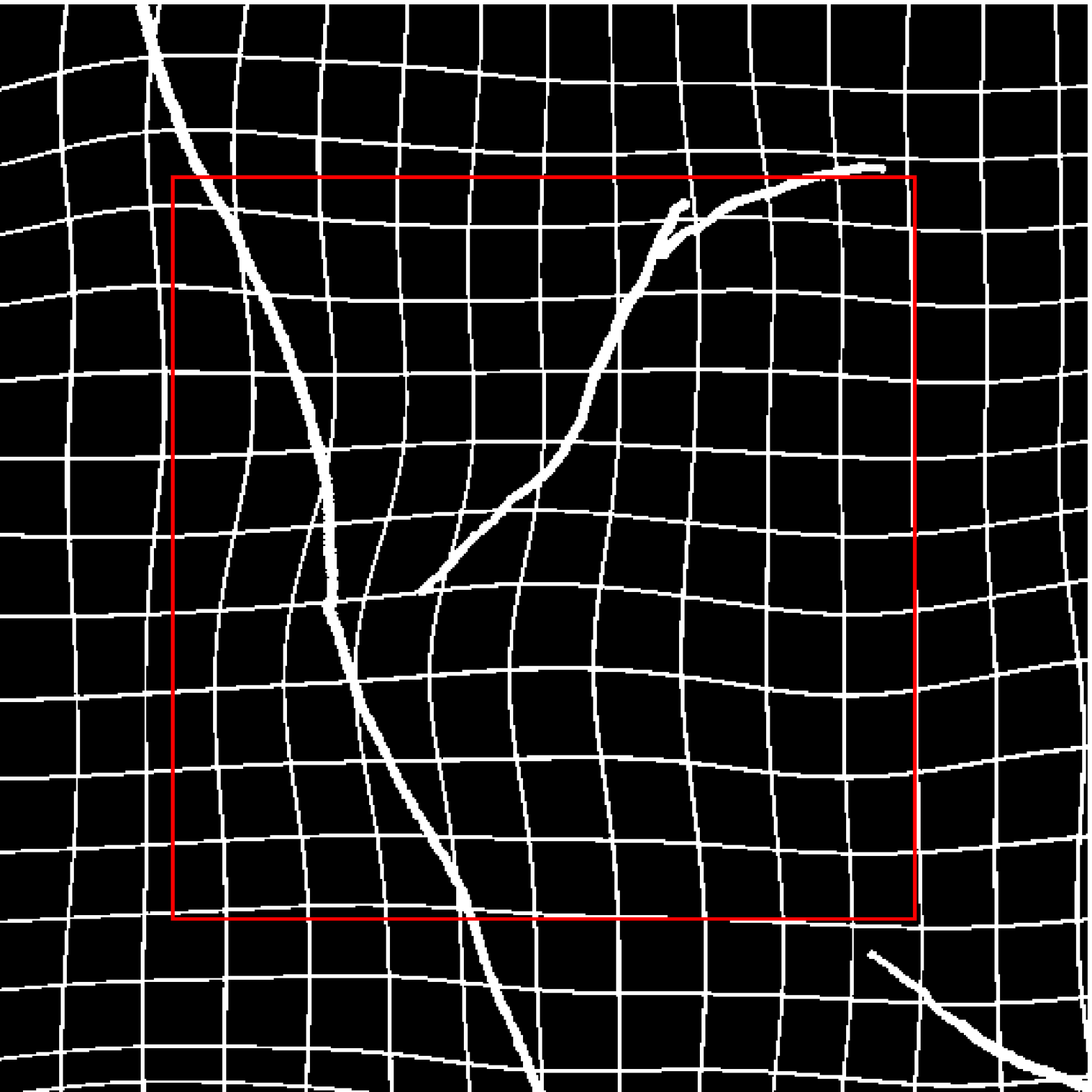}
        }
        \put(-3, 7){\huge{b}} % 10, 200 is top left
    \end{picture}
    
        \caption{(a) Elastic grid applied to an image tile 
                 and (b) corresponding annotation.
                 A white grid is shown to better illustrate the elastic grid effect.
                 A red rectangle illustrates the region which will 
                 be segmented. Augmentations such as elastic grid are designed to increase the 					 likelihood that the network will work on similar data that is not included in the training set. 				}\label{fig:aug_elastic}
    \end{figure}

    Each individual image tile was normalised to $[-0.5, +0.5]$ as centering inputs
    improves the convergence of networks trained with gradient
    descent \citep{le1991eigenvalues}.  Data augmentation is a way to artificially expand
    a dataset and has been found to improve the accuracy of CNNs for image
    classification \citep{perez2017effectiveness}.
    We used color jitter as implemented in PyTorch, with the parameters 0.3, 0.3, 0.2 and
    0.001 for brightness, contrast saturation and hue respectively.  We implemented
    elastic grid deformation (Figure \ref{fig:aug_elastic}) as described
    by \citep{simard2003best} with a probability of 0.9. Elastic grid deformations are
    parameterized by the standard deviation of a Gaussian distribution $\sigma$ which is
    an elasticity coefficient and $\alpha$ which controls the intensity of the
    deformation. As opposed to \citep{simard2003best} who suggests a constant value for
    $\sigma$ and $\alpha$, we used an intermediary parameter $\gamma$ sampled from $[0.0,
    1.0)$ uniformly.  $\gamma$ was then used as an interpolation co-efficient for both
    $\sigma$ from $[15, 60]$ and $\alpha$ from $[200, 2500]$. 
    We found by visual inspection that the appropriate $\alpha$ was larger for a larger $\sigma$. If a too large $\alpha$ was used for a given $\sigma$ then the image would look distorted in unrealistic ways. The joint interpolation of
    both $\sigma$ and $\alpha$ ensured that the maximum intensity level for a given
    elasticity coefficient would not lead to over distorted and unrealistic looking
    deformations. We further scaled $\alpha$ by a random amount from $[0.4, 1)$ so that
    less extreme deformations would also be applied. We consider the sampling of
    tiles from random locations within the larger images to provide similar benefits to
    the commonly used random cropping data augmentation procedure. The augmentations were
    ran on 8 CPU threads during the training process.

    \subsubsection{Loss}
    Loss functions quantify our level of unhappiness with the network predictions on the
    training set \citep{karpathy2016cs231n}. During training the network outputs a
    predicted segmentation for each input image. The loss function provides a way to
    measure the difference between the segmentation output by the network and the manual
    annotations. The result of the loss function is then used to update the network
    weights in order to improve its performance on the training set. We used the Dice
    loss as implemented in V-Net \citep{milletari2016v}.  Only 0.54\% of the pixels in the
    training data were roots which represents a class imbalance.  Training on imbalanced
    datasets is challenging because classifiers are typically designed to optimise overall
    accuracy which can cause minority classes to be ignored \citep{visa2005issues}.
    Experiments on CNNs in particular have shown the effect of class imbalance to be
    detrimental to performance \citep{buda2017systematic} and can cause issues with
    convergence.  The Dice loss is an effective way to handle class imbalanced datasets as
    errors for the minority class will be given more significance. For predictions $p$,
    ground truth annotation $g$, and number of pixels in an image $N$, Dice loss was
computed as:
\begin{equation}
    DL=1 - \frac{2 (p \cap g)}{p \cup g} =1 - \frac{2\sum_{i}^{N}p_{i}g_{i}}{\sum_{i}^{N}p_{i}+\sum_{i}^{N}g_{i}}
\end{equation}\label{eq:dice}
\noindent % I don't know why there was an indent. Perhaps some issue with the equation / label.
The Dice coefficient corresponds to $F_1$ when there are only two classes and ranges
from 0 to 1. It is higher for better segmentations. Thus it is subtracted from 1 to
convert it to a loss function to be minimized. We combined the Dice loss with
cross-entropy multiplied by 0.3, which was found using trial and error. This combination of loss functions was used because it provided better results than either loss function in isolation during our preliminary experiments.

\subsubsection{Optimization}

\begin{figure}
         \centering
         \includegraphics[width=0.98\linewidth]{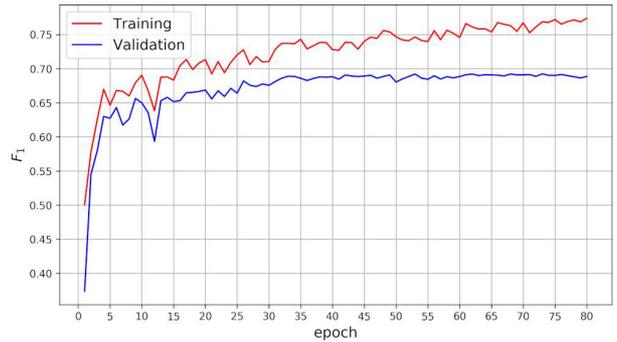}
         \caption{$F_1$ on training and validation data sets. $F_1$ is a measure of the system accuracy.
         The training $F_1$ continues to improve whilst the validation $F_1$ appears to plateau at around epoch 40.
         This is because the network is starting to fit to noise and other anomalies in the training data which are not present in the validation images.}\label{fig:unet_f1_plot} 
    \end{figure}

We used SGD with Nesterov momentum based on the formula
from \cite{sutskever2013importance}. We used a value of 0.99 for momentum as this was used in the original U-Net implementation.  We used an initial learning rate of 0.01 which was found by using trial and error whilst monitoring the validation and training $F_1$. The learning rate alters the magnitude of the updates to the network weights during each iteration of the training procedure.
    We used weight decay with a value of \num{1e-05}. A learning rate schedule was used where the learning rate
    would be multiplied by 0.3 every 30 epochs. Adaptive optimization methods such
    as Adam \citep{kingma2014adam} were avoided due to results showing they can cause
    worse generalisation behaviour
     \citep{wilson2017marginal,zhang2017yellowfin}.
    The $F_1$ computed on both the
    augmented training and validation after each epoch is shown in
    figure \ref{fig:unet_f1_plot}.

\FloatBarrier % don't let figures float outside their section
\section{Results} 
\FloatBarrier % don't let figures float outside their section

We succeeded in getting both the U-Net and the Frangi filter system to segment roots in the images in the train and validation datasets (Table \ref{table:bells3}) as well as the held out test set (Table \ref{table:test1}). As $F_1$, recall and precision is not defined for images without roots we report the results on all images combined (Table \ref{table:test1}).
%Using all images combined allows the errors in the images which didn't contain any roots to be taken into account.  The U-Net system had a higher $F_1$ than Frangi. It also had a closer balance between precision and recall. 
We report the mean and standard deviation of the per image results from the images which contain roots (Table \ref{table:test2}).
When computing these per image statistics we can see that U-Net performed better than the Frangi system for all metrics attained. 
 %As $F_1$ score is not available for the two images not containing roots, we report the total number of false positives.

\subsection{Train and validation set metrics} 

The final model parameters were selected based on the performance on the validation set. The best
validation results were attained after epoch 73 after approximately 9 hours and 34 minutes
of training. The performance on the training set was higher than the validation set (Table \ref{table:bells3}).  As parameters have been adjusted based on the data in the training and validation datasets these results are unlikely to be reliable indications of the model performance on new data so we report the performance on an unseen test set in the next section.

    \begin{table}   
    \centering
            \caption{\label{table:bells3} Best U-Net model results on the train set and the validation set used for
        early stopping. These train set results are calculated on data
        affected by both instance selection and augmentation.}
        \begin{tabular}{llll}
            & Training & Validation \\
              \toprule
            Accuracy & 0.996  & 0.996 \\
            Precision & 0.758 & 0.674 \\
            Recall & 0.780 & 0.712 \\
            $F_1$ & 0.769 & 0.692 \\
        \end{tabular}
    \end{table}

\FloatBarrier % don't let figures float outside their section
\subsection{Test set results}
\FloatBarrier % don't let figures float outside their section
\begin{table}
\centering
    \caption{\label{table:test1} Metrics on all images combined
    for the held out test set for the Frangi and U-Net segmentation systems.}
\begin{tabular}{llll}
    & Frangi & U-Net \\
    \toprule
    Accuracy            & 0.996    & 0.997 \\
    $F_1$               & 0.462    & 0.701 \\
    Precision           & 0.660   & 0.659 \\
    Recall              & 0.355   & 0.748 \\
    Prediction mean     & 0.002 & 0.006 \\
    True mean           & 0.005 & 0.005 \\

\end{tabular}

\end{table}
\begin{table}
\centering
    \caption{\label{table:test2}
    Mean and standard deviation of results on images containing roots. These are computed
    by taking the mean of the metrics computed on each of the 8 images containing roots.
    The 2 images without roots are excluded as for these $F_1$, precision and recall are
    undefined. }
\begin{tabular}{llll}
    & Frangi & U-Net \\
    \toprule
    $F_1$ mean                      & 0.463     &  0.705 \\
    $F_1$ standard deviation        & 0.085    &  0.040\\
    Recall mean                     & 0.361     &  0.749 \\
    Recall standard deviation       & 0.081    &  0.042 \\
    Precision mean                  & 0.660    &  0.666 \\
    Precision standard deviation    & 0.087    &  0.043 \\
\end{tabular}
\end{table}

\begin{figure*}
  \centering
    \begin{subfigure}[t]{0.24\textwidth}
        \centering
        \includegraphics[width=\linewidth]{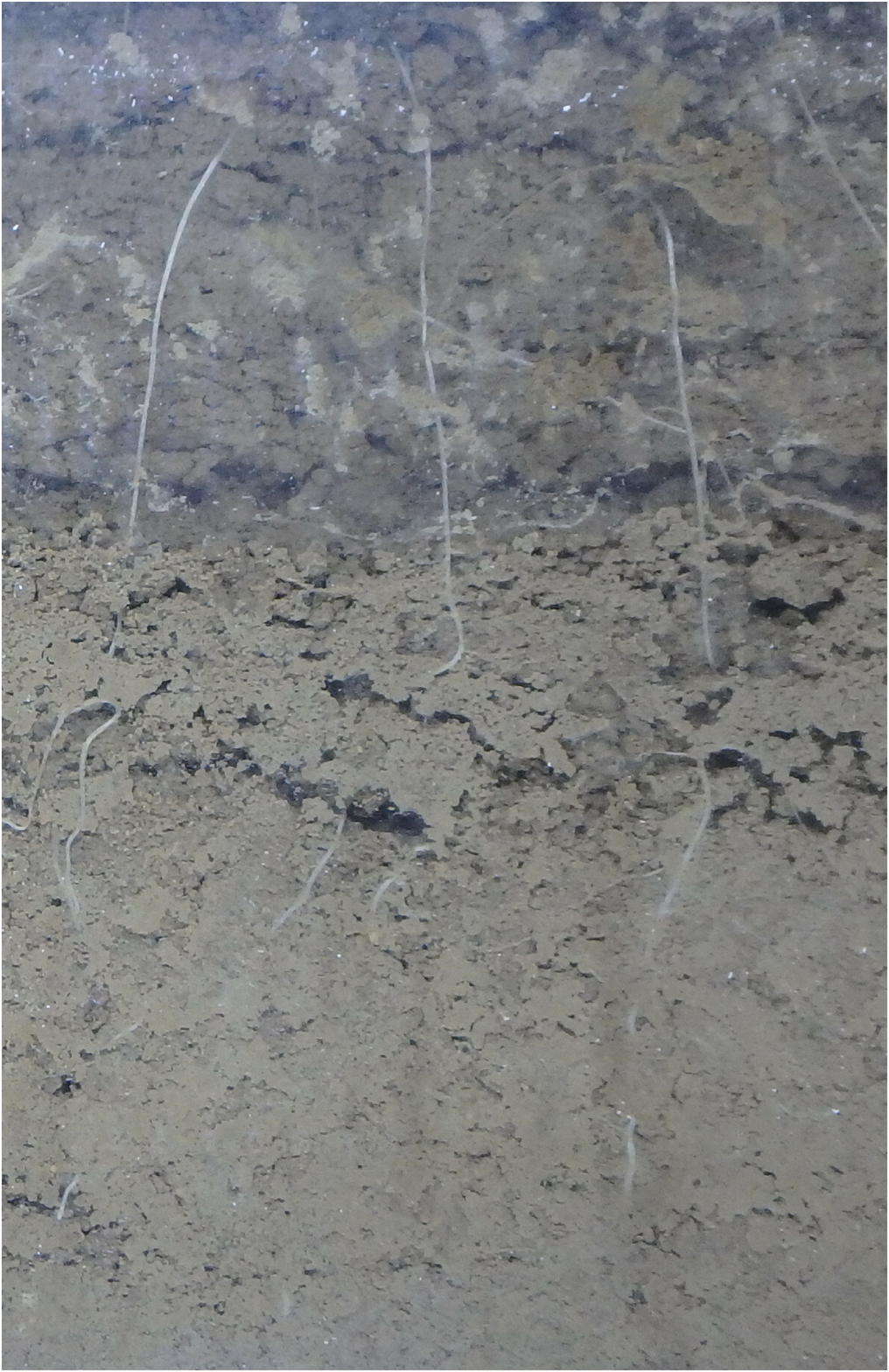}
        \caption{photo}
    \end{subfigure}
    \begin{subfigure}[t]{0.24\textwidth}
        \centering
        \includegraphics[width=\linewidth]{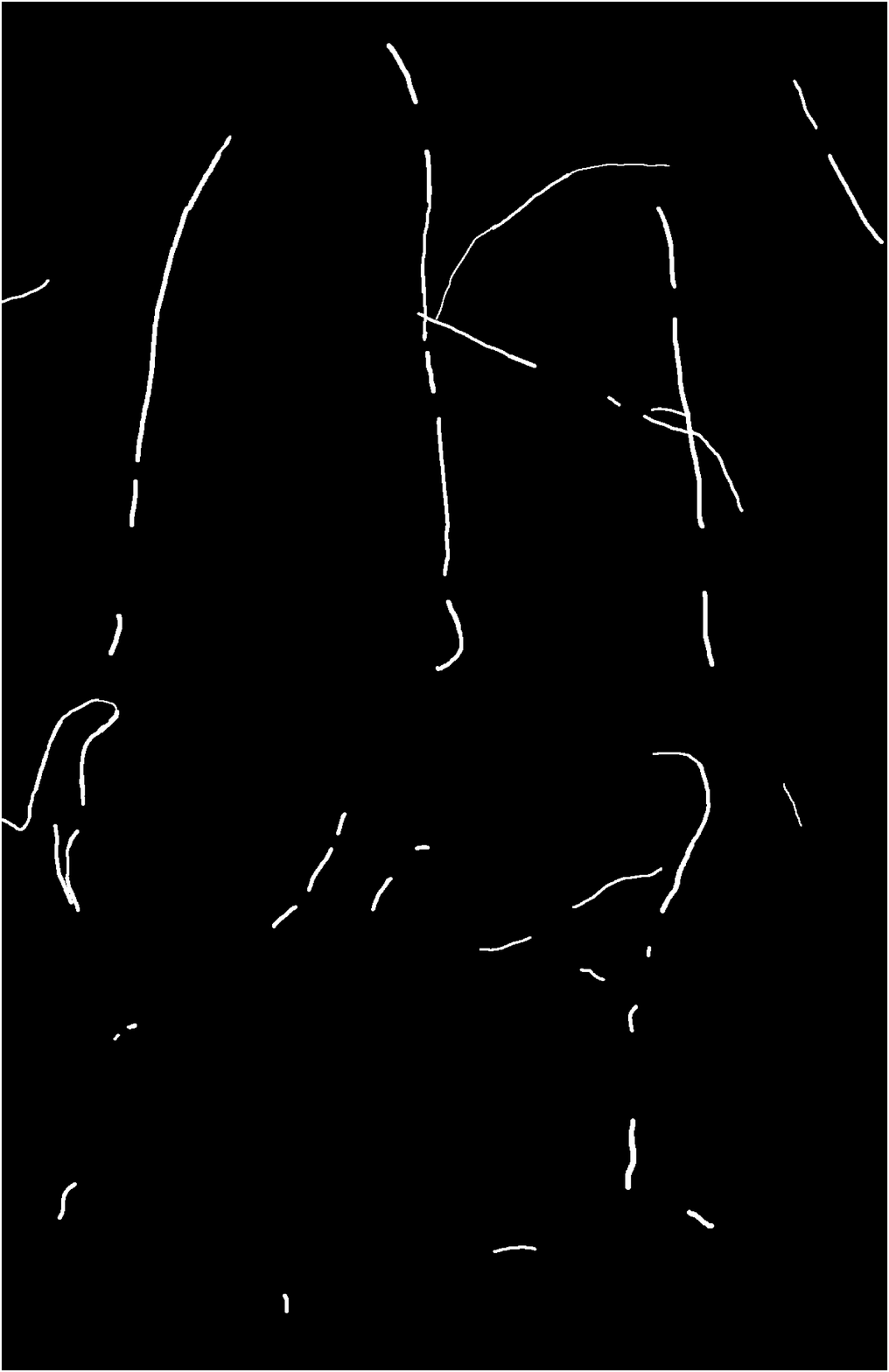}
        \caption{annotation}
    \end{subfigure}
    \begin{subfigure}[t]{0.24\textwidth}
    \centering
    \includegraphics[width=\linewidth]{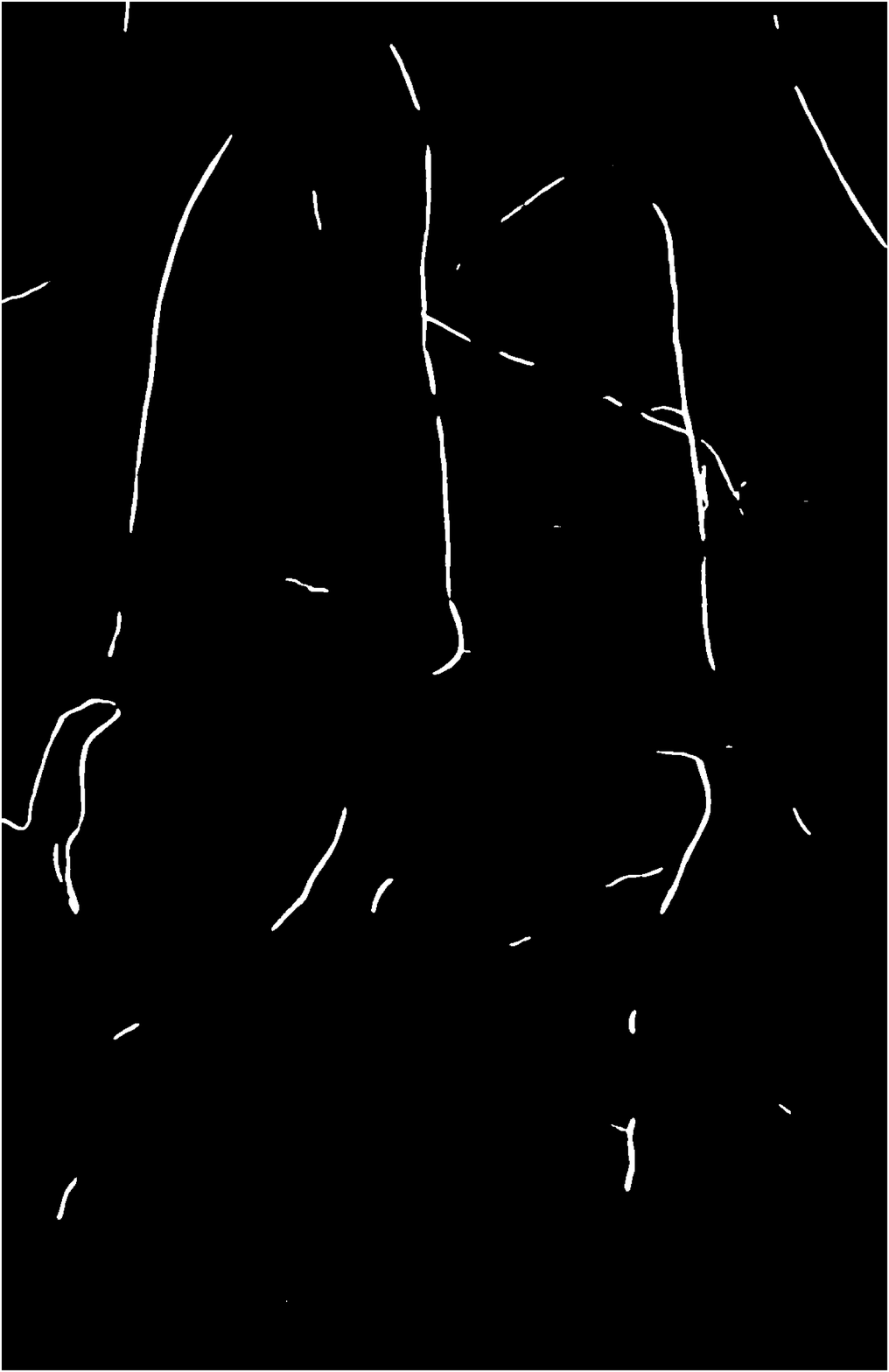}
        \caption{\label{fig:test_worst} U-Net segmentation}
    \end{subfigure}
    \begin{subfigure}[t]{0.24\textwidth}
    \centering 
    \includegraphics[width=\linewidth]{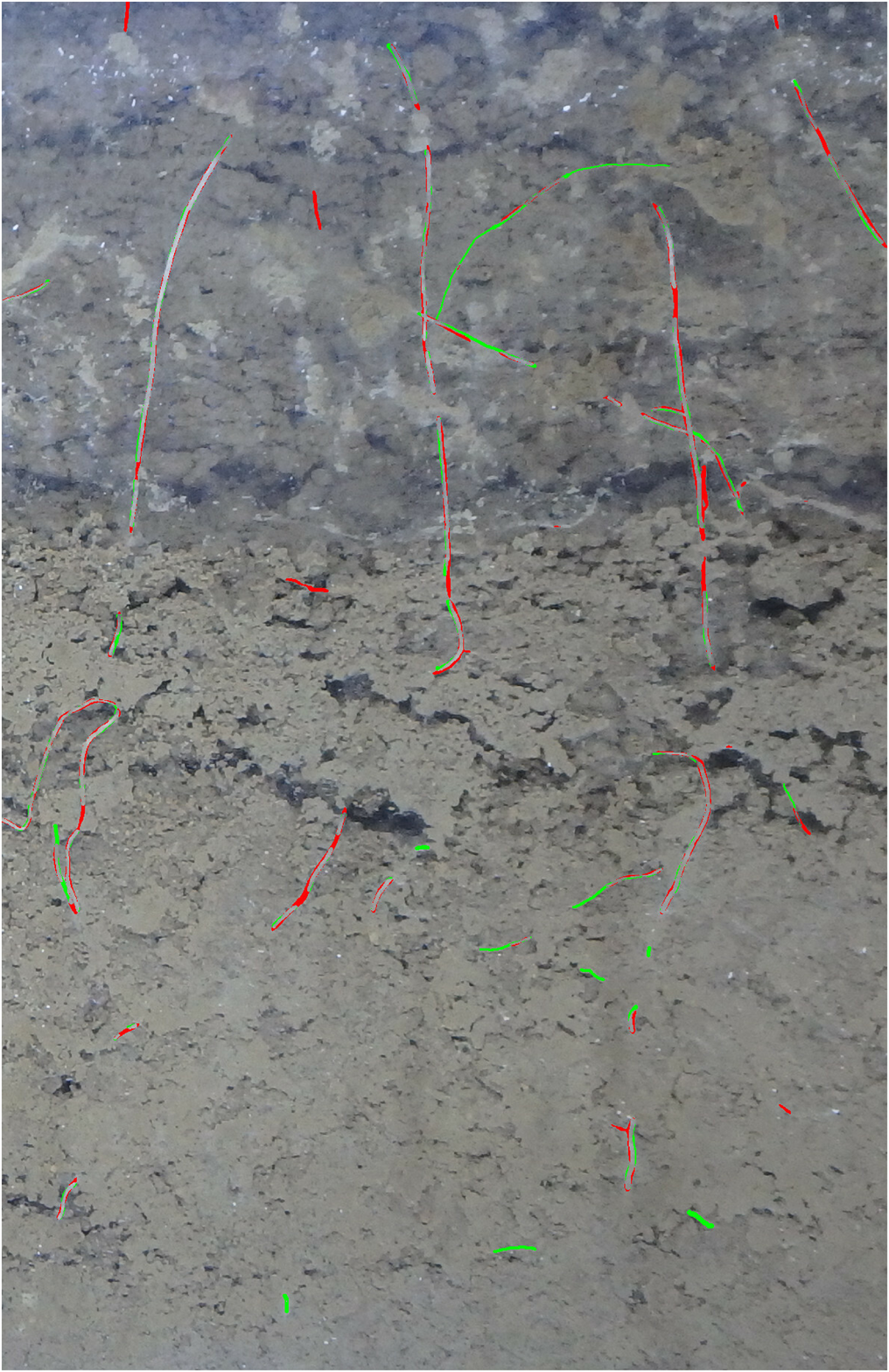}
        \caption{\label{fig:test_worst_errors} U-Net errors}
    \end{subfigure}

    \caption{\label{fig:test_worst_example} Original photo, annotation, segmentation output from U-Net and errors. To
    illustrate the errors the false positives are shown in red and the false negatives are
    shown in green. This image is a subregion of a larger image
    for which U-Net got the worst (lowest) $F_1$. }

\end{figure*}

\begin{figure*}
  \centering
    \begin{subfigure}[t]{0.24\textwidth}
        \centering
        \includegraphics[width=\linewidth]{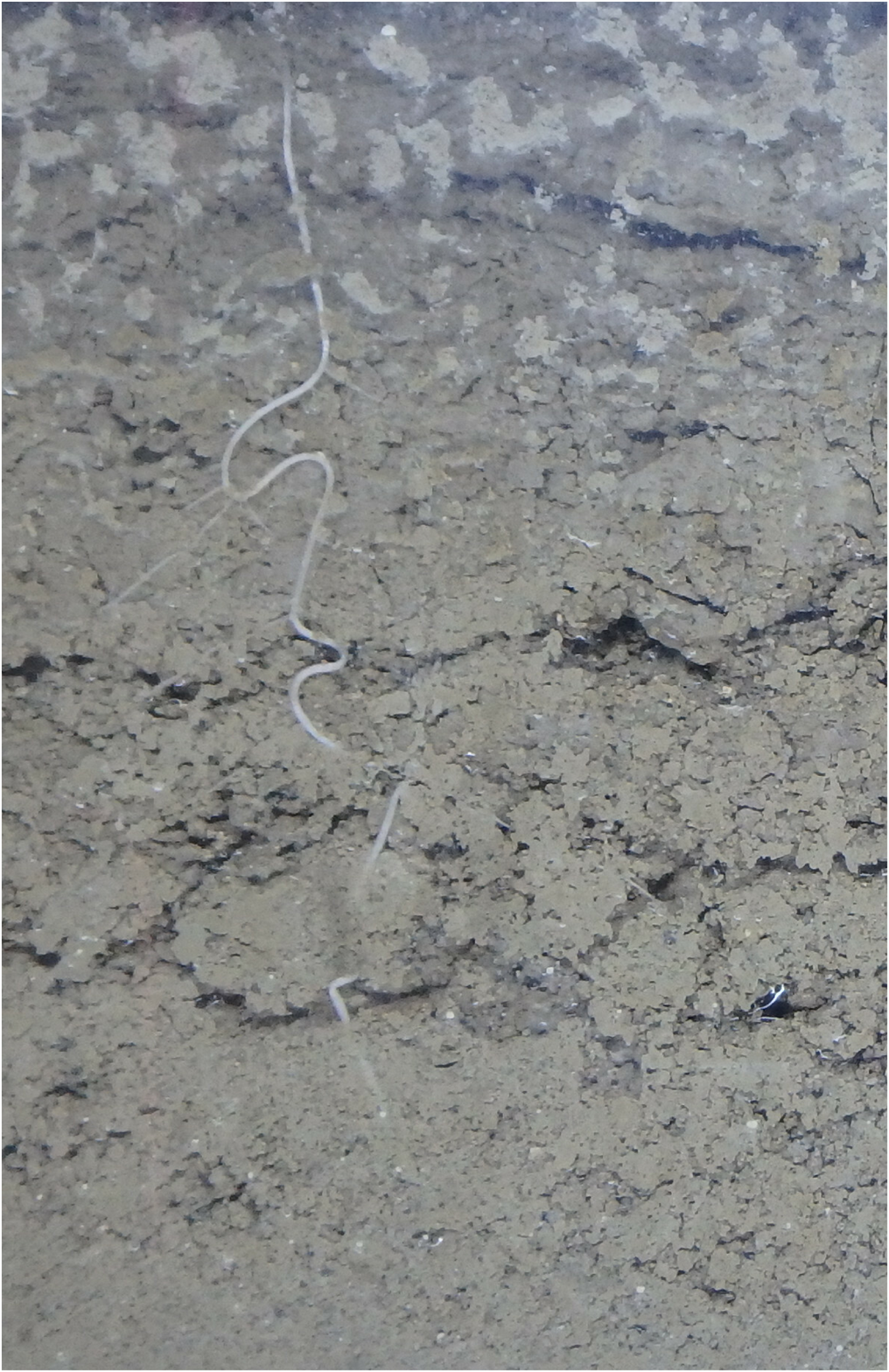}
        \caption{photo}
    \end{subfigure}
    \begin{subfigure}[t]{0.24\textwidth}
        \centering
        \includegraphics[width=\linewidth]{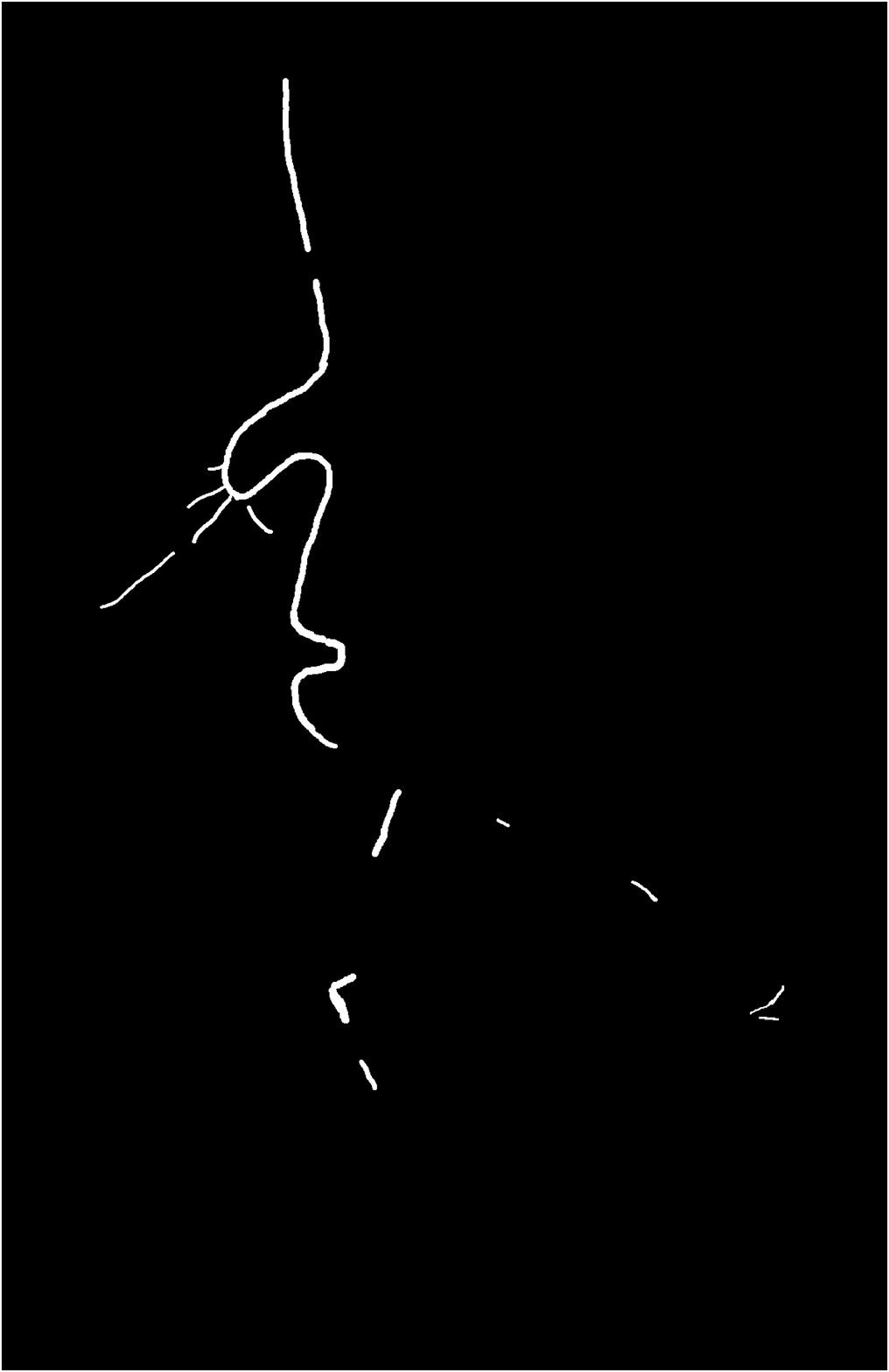}
        \caption{annotation}
    \end{subfigure}
    \begin{subfigure}[t]{0.24\textwidth}
    \centering
    \includegraphics[width=\linewidth]{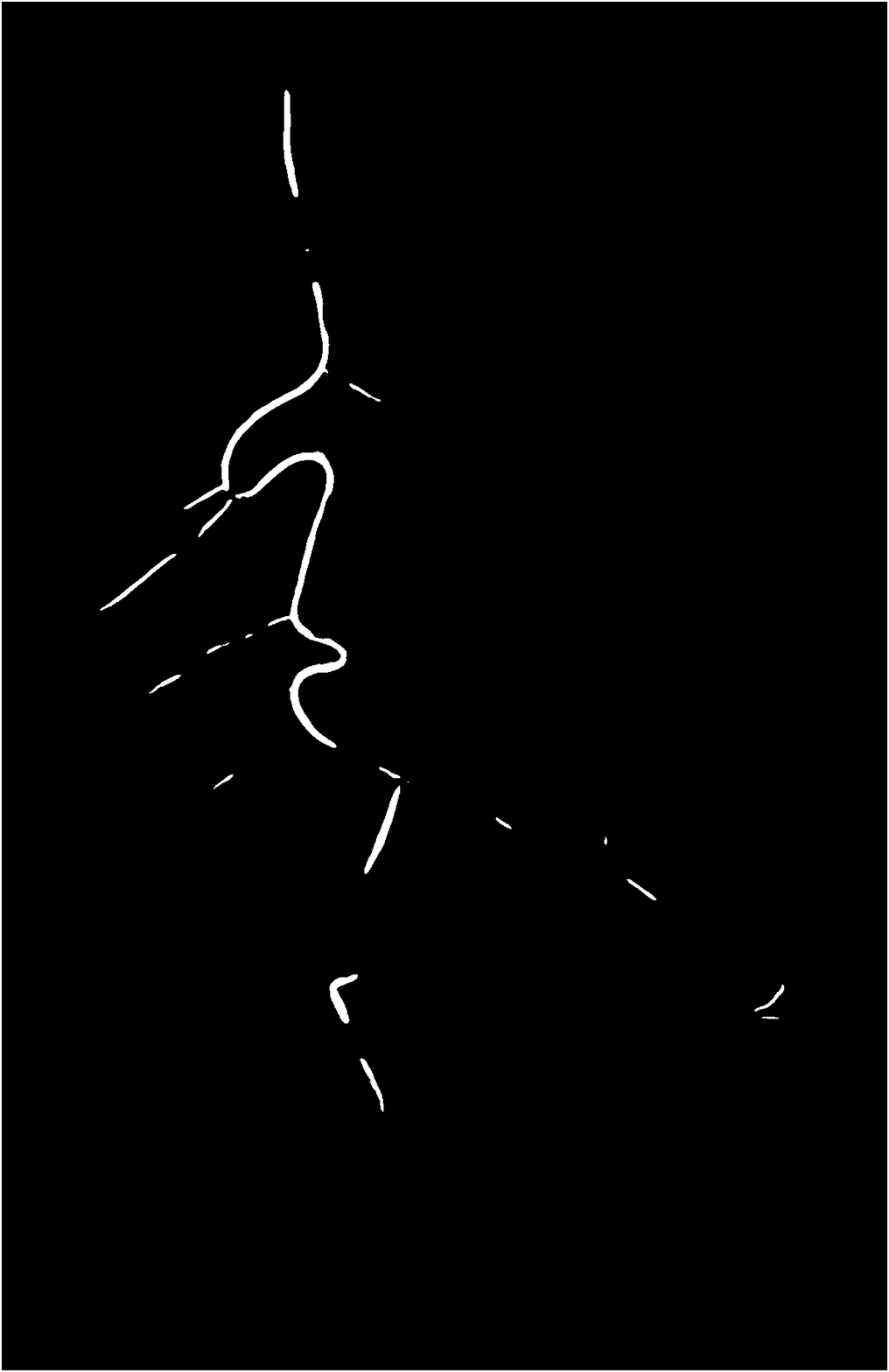}
        \caption{\label{fig:test_best} U-Net segmentation}
    \end{subfigure}
    \begin{subfigure}[t]{0.24\textwidth}
    \centering 
    \includegraphics[width=\linewidth]{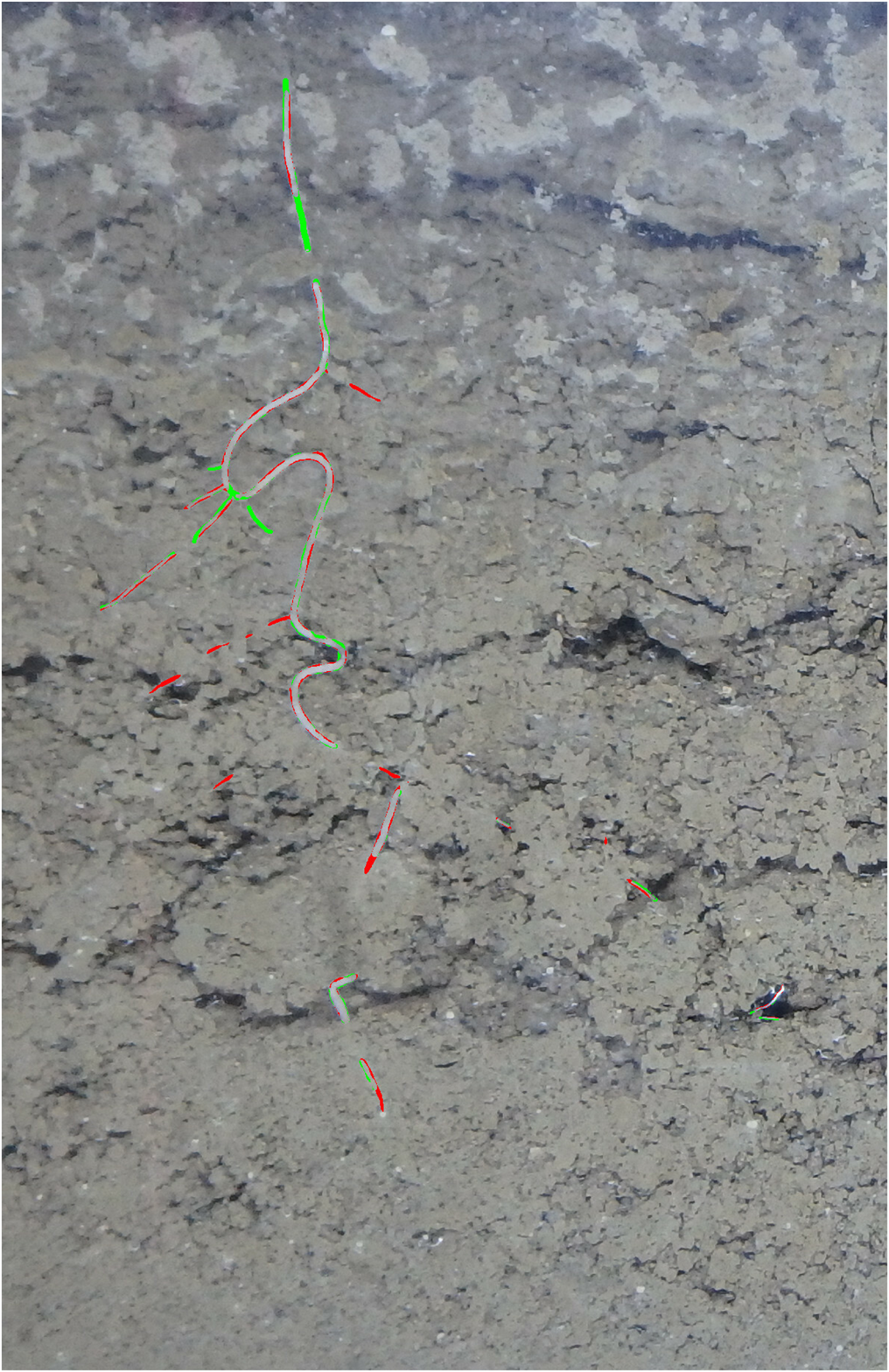}
        \caption{\label{fig:test_best_errors} U-Net errors}
    \end{subfigure}
    \caption{\label{fig:best_example}Original photo, annotation, segmentation output from U-Net and errors. To
    illustrate the errors the false positives are shown in red and the false negatives are
    shown in green. This image is a subregion of a larger image
    for which U-Net got the best (highest) $F_1$. The segmentation also contains roots which were missed by the annotator. We were able to confirm this by having the annotator review these particular errors.}
 
\end{figure*}

\begin{figure*}
 \centering
\includegraphics[width=0.16\linewidth, trim={1cm 0 0 1cm},clip]{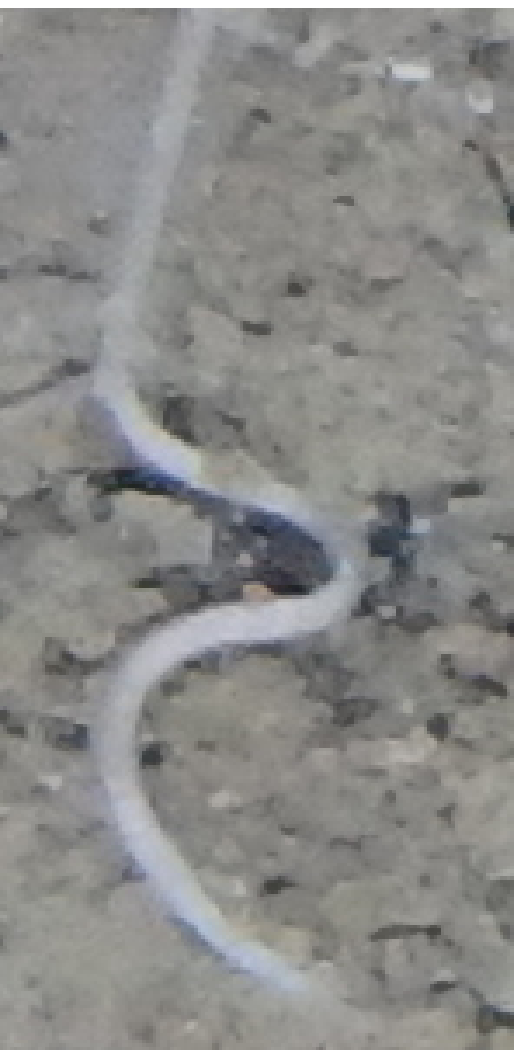}
\includegraphics[width=0.16\linewidth,trim={1cm 0 0 1cm},clip]{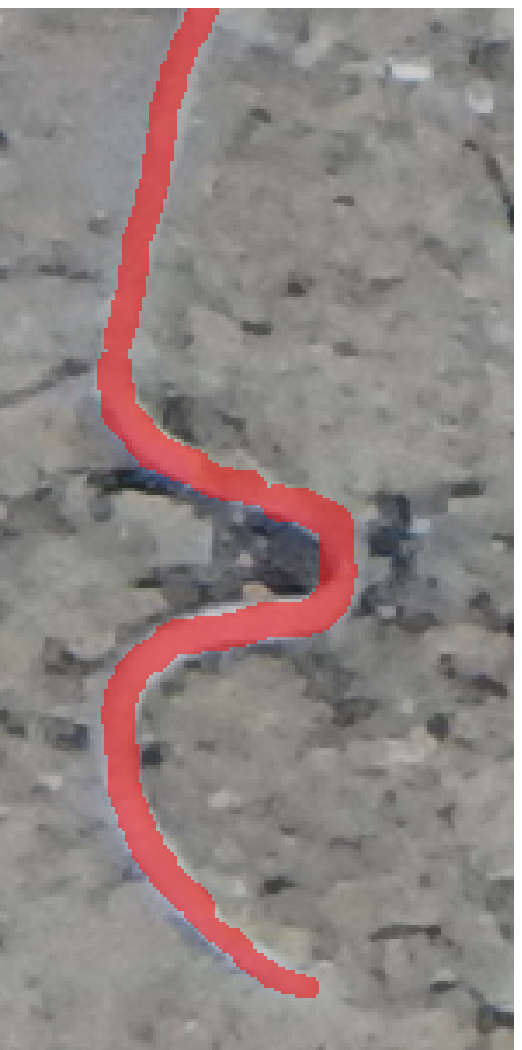}
\includegraphics[width=0.16\linewidth,trim={1cm 0 0 1cm},clip]{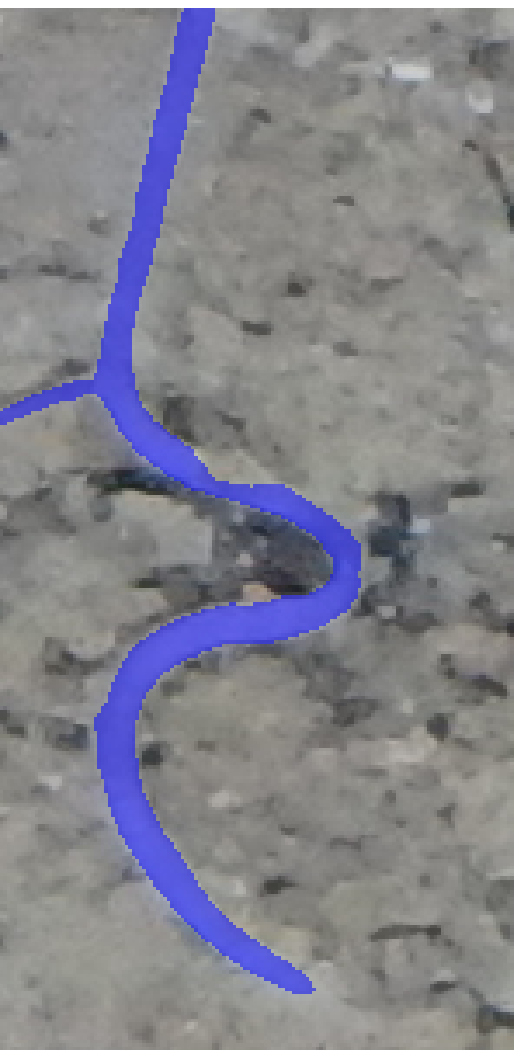}
\includegraphics[width=0.16\linewidth,trim={1cm 0 0 1cm},clip]{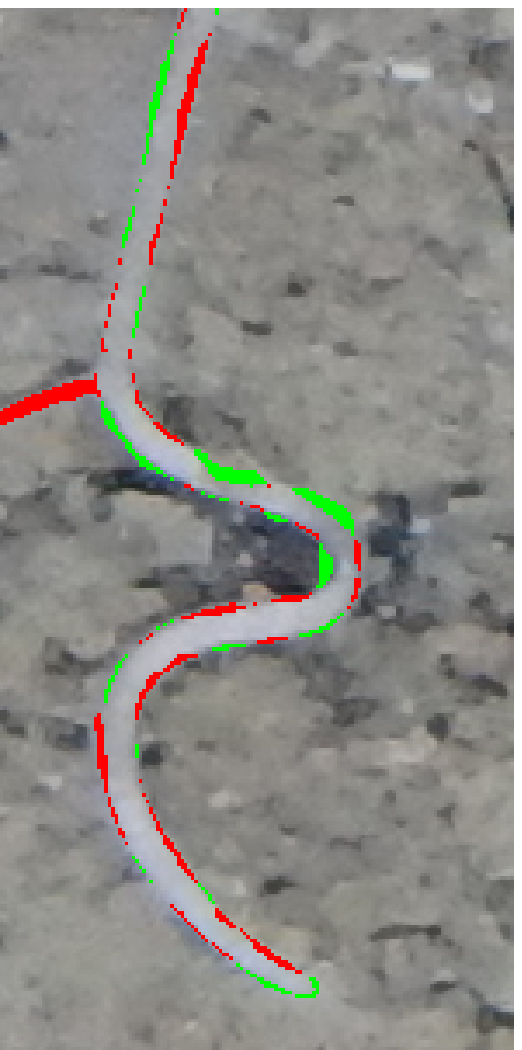}
\caption{\label{fig:discuss_annot_errors} From left to right: Image, annotation overlaid
    over image in red, U-Net segmentation overlaid over image in blue, errors with false
    positive shown in red and false negative shown in green. Many of the errors are along an ambiguous boundary region between the root and soil. Much of the error region is caused by annotation, rather than CNN segmentation errors.}
\end{figure*}

The overall percentage of root pixels in the test data was 0.49\%,
which is lower than either the training or validation dataset. Even on the image with the highest errors the CNN is able to predict many of the roots correctly 
(Figure \ref{fig:test_worst_example}). Many of the errors
appear to be on the root boundaries. Some of the fainter roots are also missed by the CNN. 
For the image with the highest (best) $F_1$ the U-Net segmentation appears very similar to the original annotation
(Figure \ref{fig:best_example}). The segmentation also contains roots which
where missed by the annotator (Figure \ref{fig:test_best_errors}) which we were able to confirm by asking the annotator to review the results. U-Net was also often able to segment the root-soil boundary more cleanly than the annotator (Figure \ref{fig:discuss_annot_errors}). False negatives can be seen at the top of the image where the CNN has failed to detect a small section of root (Figure \ref{fig:test_best_errors}).

\begin{figure*}
\centering

\begin{minipage}{.85\textwidth}
  \centering
  \includegraphics[width=1.0\linewidth]{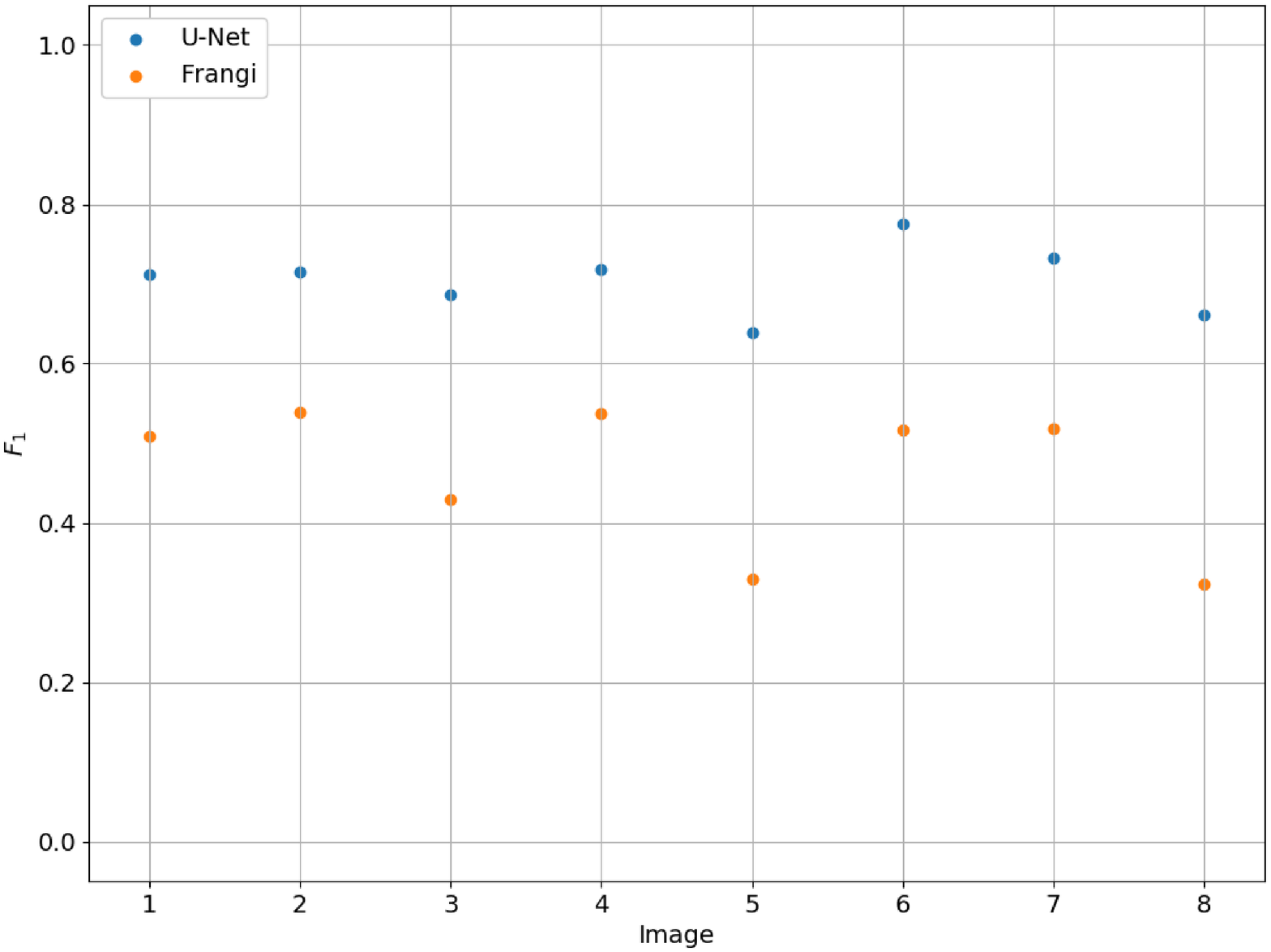}
  \captionof{figure}{The $F_1$ for the 8 images containing roots for both the Frangi and U-Net systems. }
  \label{fig:f1_all} 
\end{minipage}

\begin{minipage}{0.85\textwidth}
  \centering
     \includegraphics[width=1.0\linewidth]{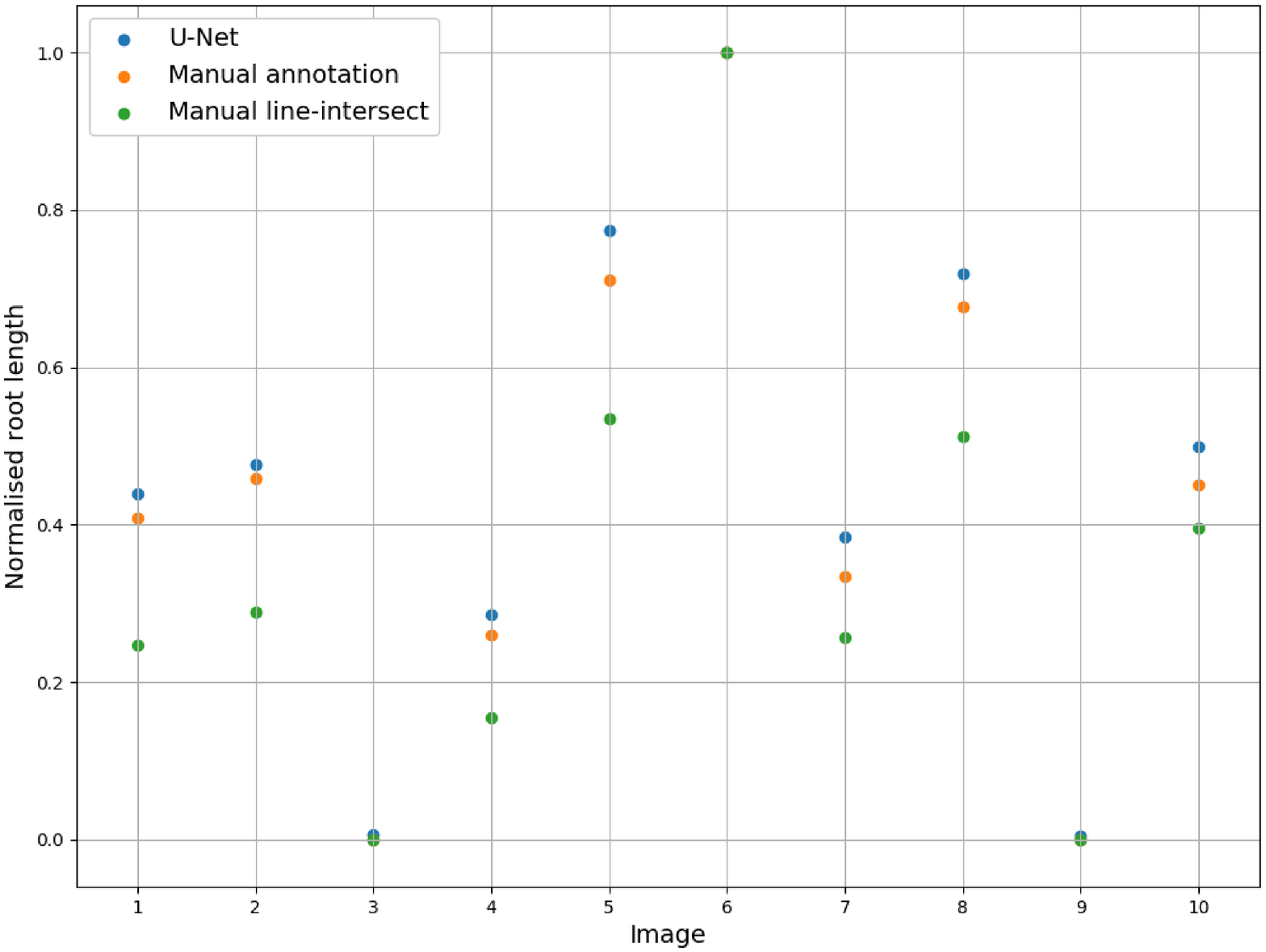}
  \captionof{figure}{Normalised root length from the U-Net segmentations, manual annotations and the line-intersect counts for the 10 test images. The measurements are normalised using the maximum value. All three methods have the same maximum value (Image 6).}
  \label{fig:ri} 
\end{minipage}
\end{figure*}

The performance of U-Net as measured by $F_1$ was better than that of the Frangi system
when computing metrics on all images combined (Table \ref{table:test1}). It also had a closer balance between precision and recall.
The U-Net segmentations have a higher $F_1$ for all images with roots in the test data (Figure \ref{fig:f1_all}). Some segmentations from the Frangi system have an $F_1$ below 0.4 whilst all the U-Net segmentations give an $F_1$ above 0.6 with the highest being just less than 0.8.
The average predicted value for U-Net was over twice that of the Frangi system. This means U-Net
predicted twice as many pixels to be root as Frangi did. 

%This difference in quantity of
% predicted root pixels is also revealed in the precision and recall. 

The slight over estimation of total root pixels
explains why recall is higher than precision for U-Net. The accuracy is above
99\% for both systems. This is because accuracy is measured as the ratio of pixels
predicted correctly and the vast majority of pixels are soil that both
systems predicted correctly.

For the two images which did not contain roots each misclassified pixel is counted as a false positive. 
The Frangi system gave 1997 and 1432 false positives on these images and the U-Net system gave 508 and 345 false positives.
The Spearman rank correlation for the corresponding U-Net
and line-intersect root intensities for the test data is 0.9848 ($p=2.288 \times 10^{-7}$).
The U-Net segmentation can be seen to give a similar root intensity to the
manual annotations (Figure \ref{fig:ri}).

\begin{figure*}
  \centering
    \includegraphics[width=1.0\linewidth]{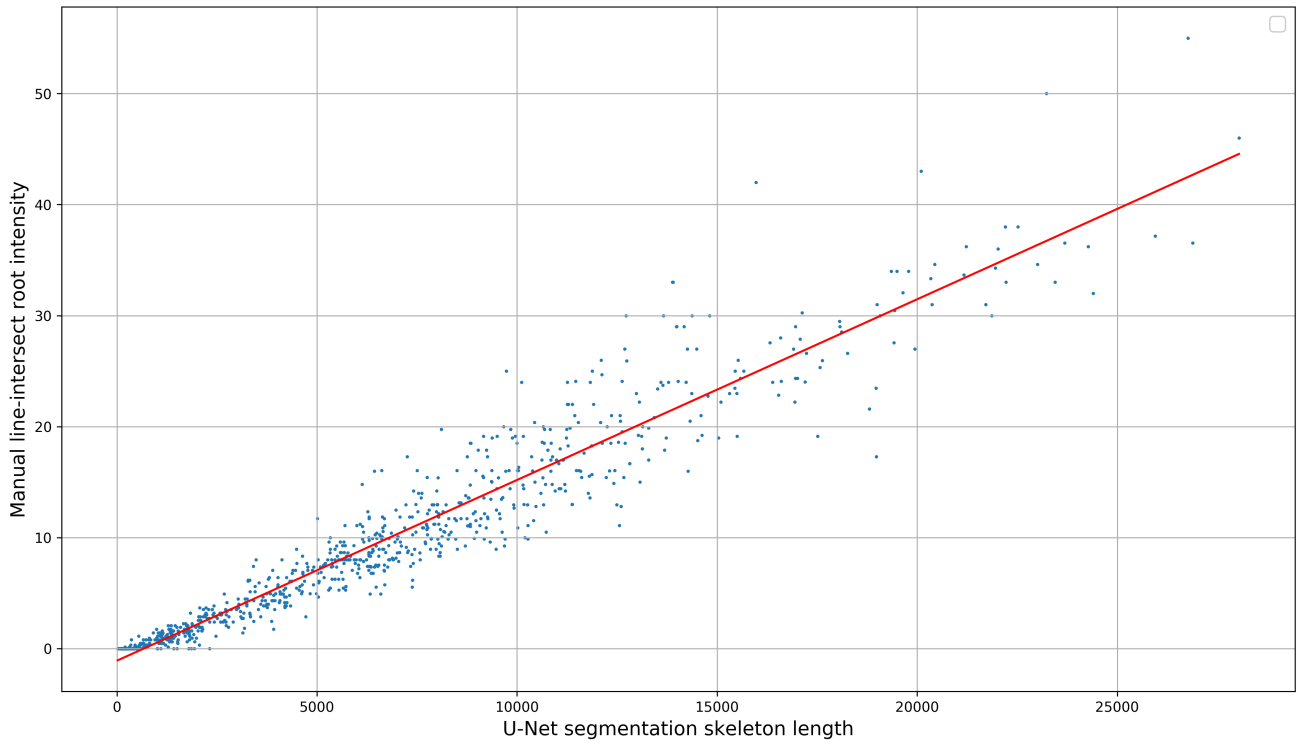}
    \caption{\label{fig:ri_vs_skel_more} RI vs segmented root length for 867 images taken in 2016. The two measurements have a Spearman rank correlation of 0.9748 %(p < $10^-8$)
    and an $R^2$ of 0.9217. 
}
\end{figure*}

We report the root intensity with the segmented root length for 867 images taken in 2016 (Figure \ref{fig:ri_vs_skel_more}). The two measurements have a Spearman rank correlation of 0.9748 $(p < 10^{-8})$
and an $r^2$ of 0.9217. Although the two measurements correlate strongly, there are some notable deviations including images for which U-Net predicted roots not observed by the manual annotator. From this scatter plot we can see that the data is heteroscedastic, forming a cone shape around the regression line with the variance increasing as root intensity increases in both measurements. 

\vspace{-10cm}

\FloatBarrier % don't let figures float outside their section
\section{Discussion}

We have presented a method to segment roots from soil using a CNN. The segmentation quality as shown
in Figures \ref{fig:test_worst} and \ref{fig:test_best} and the approximation of the root
length given by our automated method and the manual
line-intersect method for the corresponding images as shown in Figure \ref{fig:ri} and
Figure \ref{fig:ri_vs_skel_more} are a strong indication that the system works well for
the intended task of quantifying roots.

The high correlation coefficient between the measurements from the automated and manual methods supports our hypothesis that a trained U-Net is able to effectively discriminate between roots and soil in RGB
photographs. The consistently superior performance of the U-Net system
on the unseen test set over the Frangi system as measured by $F_1$ score supports our
second hypothesis that that a trained U-Net will outperform a Frangi filter based
approach.

The good generalisation behaviour and the success of the validation set at closely
approximating the test set error indicate we would likely not need as many annotations
for validation on future root datasets. As shown in Figure \ref{fig:ri_vs_skel_more} there
are some images for which U-Net predicted roots and the line-intersection count was 0.
When investigating these cases we found some false positives caused by scratches in the
acrylic glass. Such errors could be problematic as they make it hard to attain
accurate estimates of maximum rooting depth as the scratches could cause rooting depth to be
overestimated. One way to fix this would be to manually design a
dataset with more scratched panels in it in order to train U-Net not to classify them as roots.
Another possible approach would be to automatically find difficult regions of images using
an active learning approach such as \cite{yang2017suggestive} which would allow
the network to query which areas of images should be annotated based on its uncertainty.

An oft-stated limitation of CNNs is that they require large scale datasets \cite{ma2019flow} with thousands of densely labelled images \cite{roy2019squeeze} for annotation. In this study we were able to train from scratch, validate and test a CNN with only 50 images which were annotated in a few days by a single agronomist with no annotation or machine learning experience.  Our system was also designed to work with an existing photography setup using an ordinary off-the-shelf RGB camera. This makes our method more broadly accessible than methods which require a more complex multi-spectral camera system.

We used a loss function which combined Dice and cross entropy. In preliminary experiments we found
this combined loss function to be more effective than either Dice or cross entropy used in isolation. Both \cite{khened2018fully} and \cite{roy2017relaynet} found empirically that a combination of Dice and cross entropy
was effective at improving accuracy.  Although \cite{khened2018fully} claims the
combination of the loss functions is a way to yield better performance in terms of both pixel accuracy and segmentation metrics, we feel more research is needed to understand the exact benefits of such combined loss functions.

Converting from segmentation to root length was not the focus of the current study. The method we used consisted of skeletonization and then pixel counting. One limitation of this method is that it may lead to different length estimates depending on the orientation of the roots \cite{kimura1999accurate}. See \cite{kimura1999accurate} for an in depth investigation and proposed solutions.

Finding ways to improve annotation quality would also be a promising direction for further work.
Figure \ref{fig:discuss_annot_errors} shows how even a high quality segmentation will
still have a large number of errors due to issues with annotation quality. This
makes the $F_1$ given for a segmentation to not be representative of the systems' true
performance. \cite{menze2015multimodal} found significant disagreement between human raters in
segmenting tumour regions with Dice (equivalent to our $F_1$) scores between 74\% and 85\%.
We suspect a similar level of error is present in our root annotations and that improving annotation quality would improve the metrics.
Improved annotation quality would be particularly useful for the test and validation datasets as it would allow us to train the model to a higher performance.

One way to improve the quality of annotations would be to combine various annotations by different experts using a majority vote algorithm such as the one used by \cite{menze2015multimodal} although caution should be taken when implementing such methods as in some cases they can accentuate more obvious features, causing an overestimation of performance \citep{lampert2016empirical}.

It may also be worth investigating ways to reduce the weight of errors very close to the
border of an annotation, as seen in Figure \ref{fig:discuss_annot_errors}, these are often
issues with annotation quality or merely ambiguous boundary regions where a labelling of
either root or soil should not be detrimental to the $F_1$. One way to solve the problem
with misleading errors caused by ambiguous boundary regions is the approach taken
by \cite{pound2017deep} which involved having a boundary region around each area of
interest where a classification either way will not affect the overall performance
metrics.

For future research we aim to explore how well the segmentation system performance
will transfer to photographs from both other crop species and different experimental
setups. In our work so far we have explored ways to deal with a
limited dataset by using data augmentation. Transfer learning is another technique which
has been found to improve the performance of CNNs when compared to training from scratch for
small datasets \citep{tajbakhsh2016convolutional}. We can simultaneously investigate both
transfer learning and the feasibility of our system to work with different kinds of plants
 by fine-tuning our existing network on root images from new plant species.
\cite{iglovikov2018ternausnet} found pre-training U-Net to both substantially reduce
training time and prevent overfitting. Interestingly, they pre-trained U-Net on two
different datasets containing different types of images and found similar performance
improvements in both cases. Such results indicate that pre-training U-Net using images
which are substantially different from our root images may also provide performance advantages. Contra to this, \cite{he2018rethinking} found
training from scratch to give equivalent results to a transfer learning approach, which suggests that in some cases training time rather than final model performance will be the benefit of a transfer learning approach.

As opposed to U-Net, the Frangi filter is
included in popular image processing packages such as MATLAB and scikit-image.
Although the Frangi filter was initially simple to implement, we found the scikit-image
implementation too slow to facilitate optimisation on our dataset and substantial
modifications were required to make optimisation feasible.

Another disadvantage of the CNN we implemented is that as opposed to the Frangi filter, it requires a high end GPU and cannot run on a
typical laptop without further modification. \cite{mangalam2018compressing} demonstrated 
that in some cases U-Net can be compressed to 0.1\% of it's original parameter count with a very small drop in accuracy. Such an approach could be useful for making our proposed system more accessible to hardware constrained researchers.

We have demonstrated the feasibility of a U-Net based CNN system for segmenting images of roots in soil and for replacing the manual line-intersect method. The success of our approach is also a demonstration of the feasibility of deep learning in practice for small research groups needing to create their own custom labelled dataset from scratch.

\subsection{Author's Contributions}
Abraham George Smith implemented the U-Net, baseline system and wrote the manuscript with assistance from all authors.
Camilla Ruø Rasmussen did the annotations and collaborated on the introduction.
Raghavendra Selvan and Jens Petersen provided valuable machine learning expertise.

\subsection{Funding}
We thank Villum Foundation (DeepFrontier project, grant number VKR023338) for financial support for this study.

\bibliography{main}

\end{document}